\documentclass[english]{IEEEtran}
\usepackage[T1]{fontenc}
\usepackage[latin9]{inputenc}
\usepackage{xcolor}
\usepackage{pdfcolmk}
\usepackage{array}
\usepackage{float}
\usepackage{textcomp}
\usepackage{amsmath}
\usepackage{amsthm}
\usepackage{amssymb}
\usepackage{stackrel}
\usepackage{graphicx}
\usepackage{setspace}
\usepackage{wasysym}
\PassOptionsToPackage{normalem}{ulem}
\usepackage{ulem}
\makeatletter

\providecommand{\tabularnewline}{\\}
\floatstyle{ruled}
\newfloat{algorithm}{tbp}{loa}
\providecommand{\algorithmname}{Algorithm}
\floatname{algorithm}{\protect\algorithmname}
\providecolor{lyxadded}{rgb}{0,0,1}
\providecolor{lyxdeleted}{rgb}{1,0,0}

\DeclareRobustCommand{\lyxdeleted}[3]{{\color{lyxdeleted}\lyxsout{#3}}}
\DeclareRobustCommand{\lyxsout}[1]{\ifx\\#1\else\sout{#1}\fi}

\theoremstyle{plain}
\newtheorem{thm}{\protect\theoremname}
\theoremstyle{remark}
\newtheorem{rem}[thm]{\protect\remarkname}

\usepackage[caption=false, font=footnotesize]{subfig}

\usepackage{algorithmic}
\usepackage{cite}

\@ifundefined{showcaptionsetup}{}{%
 \PassOptionsToPackage{caption=false}{subfig}}
\usepackage{subfig}
\makeatother

\usepackage{babel}
\providecommand{\remarkname}{Remark}
\providecommand{\theoremname}{Theorem}

\begin{document}
\title{Single-Loop Deep Actor-Critic for Constrained Reinforcement Learning
with Provable Convergence}
\author{\singlespacing{}{\normalsize{}Kexuan Wang, An Liu, }\textit{\normalsize{}Senior Member,
IEEE}{\normalsize{} and Baishuo Lin}\thanks{Kexuan Wang, An Liu, and Baishuo Lin are with the College of Information
Science and Electronic Engineering, Zhejiang University, Hangzhou
310027, China (email: \{kexuanWang, anliu, linbaishuo\}@zju.edu.cn).
\textit{(Corresponding Author: An Liu)}}}
\maketitle
\begin{abstract}
Deep actor-critic (DAC) algorithms, which combine actor-critic with
deep neural network (DNN), have been among the most prevalent reinforcement
learning algorithms for decision-making problems in simulated environments.
However, the existing DAC algorithms are still not mature to solve
realistic problems with non-convex stochastic constraints and high
cost to interact with the environment. In this paper, we propose a
single-loop DAC (SLDAC) algorithmic framework for general constrained
reinforcement learning problems. In the actor module, the constrained
stochastic successive convex approximation (CSSCA) method is applied
to better handle the non-convex stochastic objective and constraints.
In the critic module, the critic DNNs are only updated once or a few
finite times for each iteration, which simplifies the algorithm to
a single-loop framework. Moreover, the variance of the policy gradient
estimation is reduced by reusing observations from the old policy.
The single-loop design and the observation reuse effectively reduce
the agent-environment interaction cost and computational complexity.
Despite the biased policy gradient estimation incurred by the single-loop
design and observation reuse, we prove that the SLDAC with a feasible
initial point can converge to a Karush-Kuhn-Tuker (KKT) point of the
original problem almost surely. Simulations show that the SLDAC algorithm
can achieve superior performance with much lower interaction cost.
\end{abstract}

\begin{IEEEkeywords}
Constrained/Safe reinforcement learning, deep actor-critic, theoretical
convergence.
\end{IEEEkeywords}

\section{Introduction}

\subsection{Background}

The frameworks of reinforcement learning (RL) algorithms are typically
categorized into three types: actor-only, critic-only, and actor-critic
(AC). Among them, the AC is considered to have the most potential
because it combines the strengths of the other two. Specifically,
it alternates between the critic module which estimates the state-action
value (Q value), and the actor module, which optimizes the policy
using gradients calculated from the Q values provided by the critic
module. Moreover, to handle the dimensional challenge in the case
of large state and action spaces, modern AC algorithms often parameterize
both the policy and the state-action value function (Q function) with
deep neural networks (DNNs) \cite{DRL}, namely, deep actor-critic
(DAC), which is the primary focus of this paper.

The DAC-based algorithms have achieved remarkable success in many
sequential decision-making problems, such as playing Go \cite{Go}
and Atari Games \cite{Atari}. In these simulated environments, the
agent learns to act by trial and error freely, as long as it brings
performance improvement. However, in many realistic domains such as
Question-answering systems for medical emergencies \cite{medical},
robot navigation \cite{robotlocomotion}, and radio resource management
(RRM) in future 6G wireless communications \cite{RRM}, there are
complicated stochastic constraints and high interaction cost with
the environment. For example, the RRM in wireless communications needs
to satisfy various quality of services (QoS) requirements such as
average throughput and delay, which usually involve non-convex stochastic
constraints. Moreover, each interaction with the wireless environment
involves channel estimation and/or data transmission over wireless
channels, which is quite resource-consuming. Naturally, the huge commercial
interest in deploying the RL agent to realistic domains motivates
the study of constrained reinforcement learning (CRL) with as few
agent-environment interactions as possible.

A standard and well-studied formulation for the CRL problem mentioned
above is the constrained Markov Decision Process (CMDP) \cite{SCAOPO}.
In a CMDP, the agent attempts to maximize its expected total reward
while ensuring constraints on expectations of auxiliary costs. A large
body of works have been proposed for CMDPs \cite{reviewer1_1_3,reviewer1_1_1,reviewer1_4_6,primal_dual_conv1,primal_dual_conv2,SCAOPO17,SCAOPO18,reviewer1_4_8,reviewer1_4_7,PPOLagTRPOLag},
however, among which the DAC-based algorithms are still absent and
far from mature. The classic DAC algorithms \cite{reviewer1_4_5,14in1_4_5,2019b,PPOLagTRPOLag}
for MDPs/CMDPs are generally developed under the two-loop framework,
which constantly performs the critic update to accurately estimate
the Q value for each iteration to avoid the accumulation of errors.
However, the two-loop framework, as well as the commonly adopted on-policy
sampling setting make the interaction between the agent and environment
extremely costly. Furthermore, they usually simply adopt the stochastic
(natural) gradient descent method to update policy without considering
the non-convexity. In terms of theoretical analyses, to the best of
our knowledge, no DAC-based algorithms for CMDPs have established
a strictly global convergence guarantee. In fact, for the case of
continuous action state space with policy parameterization, even other
RL algorithms that are not based on the DAC framework can rarely guarantee
the final convergence to the feasible set, except for our previous
work \cite{SCAOPO}.

To combat these weaknesses above, we propose a single-loop deep actor-critic
(SLDAC) algorithm for CRL problems in this paper, where both the policy
and the Q functions are parameterized by DNNs. In the actor module,
considering the stochasticity and the non-convexity of the objective
function and the constraints, the constrained stochastic successive
convex approximation (CSSCA) method is adopted to replace the commonly
used stochastic (natural) gradient descent in the existing DAC algorithms.
In the critic module, we use the Temporal-Difference (TD) learning
method to update the critic DNNs. To reduce the agent-environment
interaction cost and computational complexity, we perform the critic
update only once or a few finite times for each iteration, which simplifies
the algorithm to a single-loop framework (which is the so-called two-timescale
framework). In particular, we also allow the observations generated
by the old policy to be reused. Moreover, we manage to prove that
the proposed SLDAC can converge to a Karush-Kuhn-Tuker (KKT) point
of the original problem almost surely with a feasible initial point,
even with a one-step critic update in each iteration.

\subsection{Related Works}

In this section, we first review existing RL algorithms for CMDPs.
Then, since there are few works on the theoretical analysis of DAC
for CMDPs, we investigate the related works of DAC under unconstrained
cases.

\subsubsection*{RL Algorithms for CMDPs\label{subsec:Methods-for-CMDP}}

The major research lines of RL algorithms for CMDPs can be categorized
into linear programming (LP) algorithms, primal-dual algorithms, and
constrained policy optimization (CPO) algorithms based on the approach
to policy optimization \cite{CRLcategory}. By utilizing the convexity
of objective respect to state-action occupancy measure, works such
as \cite{reviewer1_1_1} and \cite{reviewer1_1_3} propose a class
of convergence-guaranteed algorithms based on LP method for CMDPs.
However, the complexity and convergence rate of the LP-based algorithms
are related to the size of the state and action space, and they are
thus not applicable to more general problems with large state and
action spaces. A common alternative of LP is the primal-dual method,
which casts the CMDPs into an unconstrained max-min saddle-point problem
and searches for the optimal policy in the primal-dual domain. However,
the common shortcoming of this category of algorithms is that they
cannot strictly guarantee the final convergence to the feasible set.
Although \cite{reviewer1_1_2} proves that the CRL problem for policies
belonging to a general distribution class can be solved exactly in
the convex dual domain, however, how to obtain the solution of non-convex
optimization in the primal domain is not analyzed. \cite{reviewer1_4_6,primal_dual_conv1,primal_dual_conv2}
only establish value-average or policy-mixture convergence, which
allows the agent to violate the safety constraints by oscillating
around an optimal safety policy. To mitigate the oscillations, \cite{SCAOPO17}
and \cite{SCAOPO18} propose a proportionally controlled Lagrangian
method and an optimistic primal-dual proximal policy optimization
algorithm, respectively. However, these two works have no theoretical
convergence guarantees. The oscillation problem is partially solved
by \cite{reviewer1_4_8} but the constraint violations still exist.
Although \cite{reviewer1_4_7} provides a theoretical result of last-iterate
convergence, it shows that the algorithm can only obtain a near-optimal
final policy in the case of finite action states space with policy
parameterization. In contrast, \cite{SCAOPO} proposes a novel successive
convex approximation-based off-policy optimization (SCAOPO) algorithm,
which addresses the CMDP problem by solving a sequence of convex objective/feasibility
optimization problems and establishes a convergence guarantee to a
KKT point, which greatly inspires this paper. However, the above works
are not based on the DAC framework. Recently, authors in \cite{PPOLagTRPOLag}
connect the primal-dual method to two-loop DAC and propose the trust
region policy optimization Lagrangian (TRPO-Lag) and proximal policy
optimization Lagrangian (PPO-Lag), however, the global convergence
of which is unknown. In addition, authors in \cite{CPO} propose a
CPO algorithm, which updates the policy within a trust region, but
it also cannot theoretically assure a strictly feasible result. 

Moreover, reinforcement learning algorithms for CMDPs can also be
divided into online CRL \cite{boundedJ2} and offline CRL \cite{offlineCRLrecommendedbyreviewer2}
from the perspective of data acquisition, where the former aims to
learn a task-solving policy by interacting with the environment, while
the later chooses to learn from offline datasets. Due to the discrepancy
between the offline dataset and the real-world environment, the offline
CRL is usually only suited for the pre-training stage. Therefore,
we mainly focus on the online setting in this paper, while leaving
the research of the offline setting in our future work.

\subsubsection*{Convergence Analysis for Deep Actor-Critic}

A large body of works establishes the convergence guarantee for DAC
in the unconstrained case, i.e., \cite{reviewer1_4_5,14in1_4_5,2019b}
for two-loop DAC and \cite{reviewer1_4_3,9in1_4_4,11in1_4_4} for
the so-called single-loop (two-timescale) DAC. Recently, authors in
\cite{reviewer1_4_4} propose the single-timescale DAC, which further
proves that the actor module and the critic module can converge on
the same timescale. However, all the above DAC algorithms simply adopt
the stochastic (natural) policy descent, which makes them only suitable
for simple constraints with deterministic convex feasible sets. In
addition, the challenging observation reuse setting is rarely considered.

\subsection{Contributions}

The contributions in this paper are summarized below:
\begin{itemize}
\item \textbf{Non-convex Constrained Optimization and Low-cost Settings:}
Compared with the existing DAC algorithms, the proposed SLDAC pays
more attention to addressing the requirements of CRL problems in the
realistic domains. There are two major differences/advantages: 1)
our method takes into account the stochasticity and the non-convexity
of both the objective function and the constraints; 2) the combination
of single-loop framework and the observation reuse can significantly
reduce the agent-environment interaction cost and computational complexity.
\item \textbf{Theoretical Convergence Analysis of SLDAC:} Under some technical
conditions, we prove that the proposed SLDAC can converge to KKT points
of the original CMDP problem with a feasible initial point, despite
the biased estimation induced by the observation reuse and the single-loop
framework. To the best of our knowledge, it is the first single-loop
DAC algorithm for CMDPs that can provably converge to a KKT point.
Moreover, we also provide some valuable theoretical results, including
the asymptotic consistency of the estimated function values and policy
gradients, as well as the finite-time convergence rates of both the
critic DNNs. 
\end{itemize}
The rest of the paper is organized as follows. Section II briefly
introduces some preliminaries, including general formulations of CRL,
neural network parameterization, and assumptions on problem structure.
The SLDAC algorithmic framework and its convergence analysis are presented
in Sections III and IV, respectively. Section V provides simulation
results and the conclusion is drawn in section VII.

\section{Preliminaries\label{sec:Preliminaries}}

In this section, we first introduce some preliminaries of CMDP and
then briefly introduce a family of DNNs, which are commonly used in
modern algorithms to parameterize the policy $\pi$ and Q functions.
Moreover, we make several assumptions on the problem structure, before
we present the algorithmic framework.

\subsection{Problem Formulation\label{subsec:Problem-Formulation}}

A CMDP can be denoted as a tuple $\left(\mathcal{S},\mathcal{\mathcal{A}},P,C\right)$,
where $\mathcal{S}\subseteq\mathbb{R}^{n_{\boldsymbol{s}}}$ is the
state space, $\mathcal{\mathcal{A}}\subseteq\mathbb{R}^{n_{\boldsymbol{a}}}$
is the action space, $P:\mathcal{S}\times\mathcal{A}\times\mathcal{S}\rightarrow\left[0,1\right]$
is the transition probability function, where $P\left(\boldsymbol{s}'\mid\boldsymbol{s},\boldsymbol{a}\right)$
denotes the probability of transition to state $\boldsymbol{s}'$
from state $\boldsymbol{s}\in\mathcal{S}$ with an action $\boldsymbol{a}\in\mathcal{\mathcal{A}}$,
and $C_{i=0,1,\ldots}:\mathcal{S}\times\mathcal{A}\rightarrow\mathbb{R}$
are the per-stage reward/cost functions. The policy $\pi:\mathcal{S}\rightarrow\mathbf{P}\left(\mathcal{A}\right)$
is a map from states to an action probability distribution, with $\pi\left(\boldsymbol{a}\mid\boldsymbol{s}\right)$
denoting the probability of selecting action $\boldsymbol{a}\in\mathcal{A}$
in state $\boldsymbol{s}\in\mathcal{S}$. Then, the transition probability
$P$ and policy $\pi$ together determine the probability distribution
of the trajectory $\left\{ \boldsymbol{s}_{0},\boldsymbol{a}_{0},\boldsymbol{s}_{1},\ldots\right\} $,
where $\boldsymbol{s}_{t}$ and $\boldsymbol{a}_{t}$ denote the state
and action at time step $t$. For simplicity, we denote the probability
distribution over the trajectory by $p_{\pi}$, i.e., $\boldsymbol{s}_{t}\thicksim P\left(\cdot\mid\boldsymbol{s}_{t-1},\boldsymbol{a}_{t-1}\right),\boldsymbol{a}_{t}\thicksim\pi\left(\cdot\mid\boldsymbol{s}_{t}\right)$.

Because of the curse of dimension, we parameterize the policy $\pi$
with DNN over $\boldsymbol{\theta}\in\mathbf{\Theta}$, and then we
denote the parameterized policy as $\pi_{\boldsymbol{\theta}}$. With
the $\pi_{\boldsymbol{\theta}}$, the goal of a general CRL problem
for continuous control can be formulated based on the CMDP as:
\begin{align}
\underset{\theta\in\Theta}{\mathrm{min}}J_{0}\left(\boldsymbol{\theta}\right)\overset{\triangle}{=} & \underset{T\rightarrow\infty}{\mathrm{lim}}\frac{1}{T}\mathbb{E}_{p_{\pi_{\boldsymbol{\theta}}}}\left[\stackrel[t=0]{T-1}{\sum}C_{0}\left(\boldsymbol{s}_{t},\boldsymbol{a}_{t}\right)\right]\label{eq:problem 1}\\
\mathrm{s.t.}J_{i}\left(\boldsymbol{\theta}\right)\overset{\triangle}{=} & \underset{T\rightarrow\infty}{\mathrm{lim}}\frac{1}{T}\mathbb{E}_{p_{\pi_{\boldsymbol{\theta}}}}\left[\stackrel[t=0]{T-1}{\sum}C_{i}\left(\boldsymbol{s}_{t},\boldsymbol{a}_{t}\right)\right]-c_{i}\leq0.\nonumber 
\end{align}
where $c_{1},\ldots,c_{I}$ denote the constraint values. Problem
(\ref{eq:problem 1}) embraces many important applications, which
are specifically exemplified in the following.

\paragraph*{Example 1 (Delay-Constrained Power Control for Downlink MU-MIMO system)}

Some mission-critical applications in the forthcoming 6G communication,
such as Internet-of-Things and virtual reality applications, are emerging
and call for low-delay communication services for each user \cite{URLLC2,XR,URLLC3}.
Now we consider a delay-constrained power control problem, which aims
to obtain a policy $\pi_{\boldsymbol{\theta}}$ to satisfy this stringent
requirement with minimal long-term average power consumption. Consider
a downlink MU-MIMO system consisting of a $N_{t}$-antennas base station
(BS) and $K$ single-antenna users ($N_{t}\geq K$), where the BS
maintains $K$ dynamic data queues for the burst traffic flows to
each user. Suppose that the time dimension is partitioned into decision
slots indexed by $t$ with slot duration $t_{0}$, and each queue
dynamic of the $k$-user has a random arrival data rate $A_{k}\left(t\right)$,
where $E\left[A_{k}\right]=\lambda_{k}$. The data rate $R_{k}\left(t\right)$
of user $k$ is given by
\[
R_{k}=B\mathrm{log}_{2}\biggl(1+\frac{P_{k}\bigl|\boldsymbol{h}_{k}^{H}\boldsymbol{v}_{k}\left(\alpha_{Z}\right)\bigr|^{2}}{\sum_{j\neq k}P_{j}\bigl|\boldsymbol{h}_{k}^{H}\boldsymbol{v}_{j}\left(\alpha_{Z}\right)\bigr|^{2}+\sigma_{k}^{2}}\biggr),
\]
where we omit the time slot index for conciseness, $B$ denotes the
bandwidth, $\boldsymbol{h}_{k}$ is the downlink channel of the $k$-th
user, $\sigma_{k}^{2}$ is the noise power at the $k$-th user, $P_{k}$
is the power allocated to the $k$-th user, and $\boldsymbol{v}_{k}\left(\alpha_{Z}\right)$
is the normalized regularized zero-forcing (RZF) precoder with regularization
factor $\alpha_{Z}$ \cite{RZF}. Then, the queue dynamic of the $k$-user
$Q_{k}\left(t\right)$ is given by
\[
Q_{k}\left(t\right)=\mathrm{max}\Bigl\{ A_{k}\left(t\right)t_{0}-R_{k}\left(t\right)t_{0}+Q_{k}\left(t-1\right),0\Bigr\}.
\]
Finally, the delay-constrained power control problem can be formulated
as
\begin{align}
\underset{\theta\in\Theta}{\mathrm{min}}J_{0}\left(\boldsymbol{\theta}\right)\overset{\triangle}{=} & \underset{T\rightarrow\infty}{\mathrm{lim}}\frac{1}{T}\mathbb{E}_{p_{\pi_{\boldsymbol{\theta}}}}\left[\stackrel[t=1]{T}{\sum}\stackrel[k=1]{K}{\sum}P_{k}\left(t\right)\right]\label{eq:delay-constrained power control problem}\\
\mathrm{s.t.}J_{i}\left(\boldsymbol{\theta}\right)\overset{\triangle}{=} & \underset{T\rightarrow\infty}{\mathrm{lim}}\frac{1}{T}\mathbb{E}_{p_{\pi_{\boldsymbol{\theta}}}}\left[\stackrel[t=1]{T}{\sum}\frac{Q_{k}\left(t\right)}{\lambda_{k}}\right]-c_{k}\leq0,\forall k,\nonumber 
\end{align}
where $c_{1},\ldots,c_{K}$ denote the maximum allowable average delay
for each user. In this case, the BS at the $t$-th time slot obtains
state information $\boldsymbol{s}_{t}=\left\{ \mathbf{Q}\left(t\right),\mathbf{H}\left(t\right)\right\} $,
where $\mathbf{Q}=\left[Q_{1},\ldots,Q_{K}\right]^{\top}\in\mathbb{R}^{K}$
and $\mathbf{H}=\left[\boldsymbol{h}_{1},\ldots,\boldsymbol{h}_{K}\right]^{H}\in C^{K\times N_{t}}$,
and takes action $\boldsymbol{a}_{t}=\left\{ \boldsymbol{P}_{t}\left(t\right),\alpha_{Z}\left(t\right)\right\} $
according to policy $\pi_{\boldsymbol{\theta}}$, where $\boldsymbol{P}_{t}=\left[P_{1},\ldots,P_{K}\right]$.
Moreover, the reward function $C_{0}\left(\boldsymbol{s}_{t},\boldsymbol{a}_{t}\right)=\stackrel[k=1]{K}{\sum}P_{k}\left(t\right)$,
and the cost function $C_{1}\left(\boldsymbol{s}_{t},\boldsymbol{a}_{t}\right)=\frac{Q_{k}\left(t\right)}{\lambda_{k}}$.

\paragraph*{Example 2 (Autonomous Vehicle Transport with Safety Assurance)}

\textcolor{blue}{}
\begin{figure}
\begin{centering}
\includegraphics[width=5.3cm,height=3.2cm]{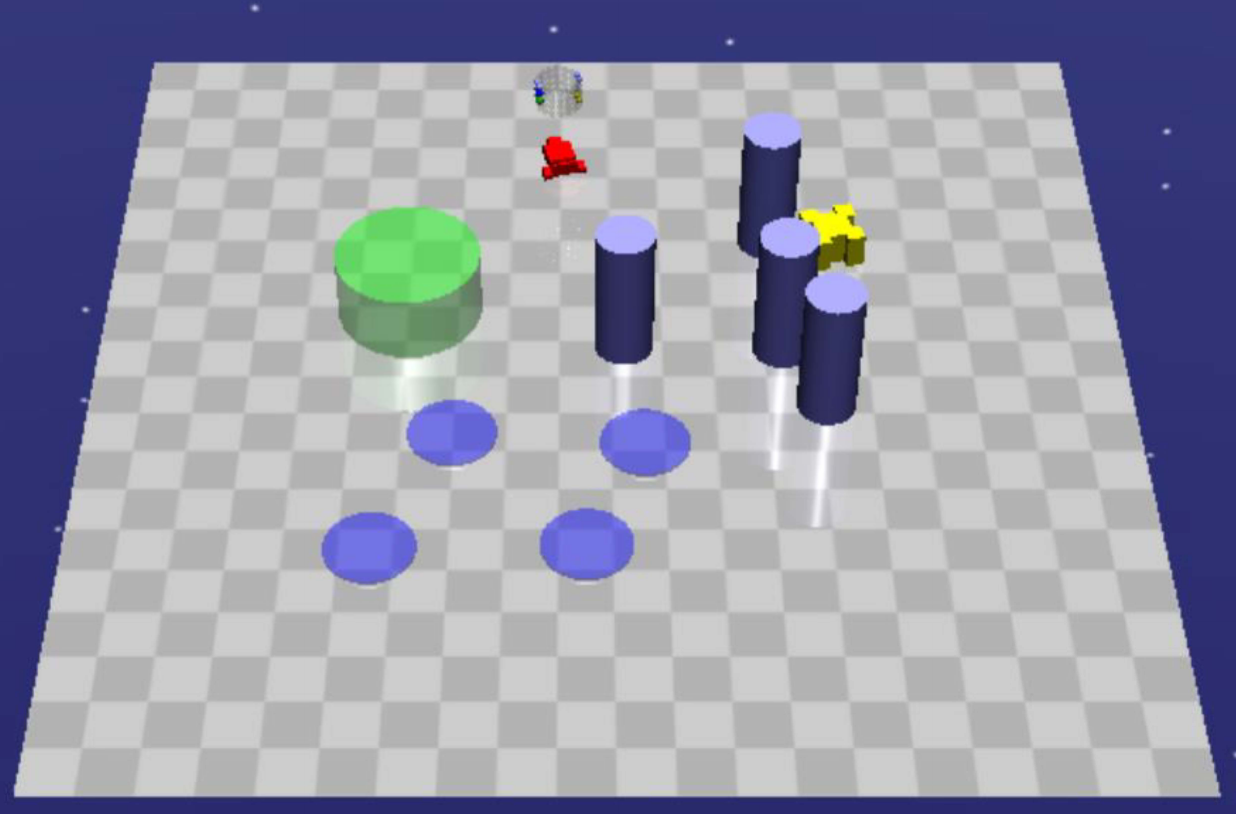}
\par\end{centering}
\caption{\textcolor{blue}{\label{fig:safety_gym}}An autonomous vehicle transport
environment provided by the Safety Gym.}
\end{figure}
Autonomous vehicles with safety assurance have attracted significant
attention in recent years. To prepare for its application in the real
world, the OpenAI Safety Gym \cite{PPOLagTRPOLag} provides a widely
used benchmarking simulated environment, as shown in Fig. \ref{fig:Safety Gym}.
A car robot with two independently driven parallel wheels and a free-rolling
rear wheel is required to find the yellow box and transport it to
the green destination while avoiding hazards (including 4 blue areas
representing primary hazards and 4 blue pillars representing) as much
as possible. In this case, the current state $\boldsymbol{s}_{t}$
is obtained by onboard radar and includes the distance and direction
information of surrounding objects and hazardous areas. The action
$\boldsymbol{a}_{t}$ is adopted to control the two independently
driven parallel wheels of the car. The reward $C_{0}\bigl(\boldsymbol{s}_{t},\boldsymbol{a}_{t}\bigr)$
obtained by the robot consists of a small, dense component encouraging
movement toward the box and the target goal and a large, sparse component
for successfully completing the task. Conversely, each time the car
enters a blue zone, it receives a cost of $C_{1}\bigl(\boldsymbol{s}_{t},\boldsymbol{a}_{t}\bigr)$=1,
and each time it touches a blue pillar, it receives a cost of $C_{1}\bigl(\boldsymbol{s}_{t},\boldsymbol{a}_{t}\bigr)$=10.
Moreover, the constraint for this problem is that the average cost
is less than or equal to $c_{1}$. 

\paragraph*{Example 3 (Constrained Linear-Quadratic Regulator)}

The constrained Linear-quadratic regulator (CLQR) is one of the most
fundamental problems in control theory \cite{LQRimportance}. According
to \cite{LQR} and \cite{SCAOPO}, by denoting $\boldsymbol{s}_{t}\in\mathbb{R}^{n_{\boldsymbol{s}}}$
and $\boldsymbol{a}_{t}\in\mathbb{R}^{n_{\boldsymbol{a}}}$ as the
state and action of CLQR problem at the time $t$, the objective cost
and the constraint cost in the CLQR setting are respectively given
by
\begin{align*}
C_{0}\left(\boldsymbol{s}_{t},\boldsymbol{a}_{t}\right) & =\boldsymbol{s}_{t}^{\mathrm{T}}Q_{0}\boldsymbol{s}_{t}+\boldsymbol{a}_{t}^{\mathrm{T}}R_{0}\boldsymbol{a}_{t},
\end{align*}
\[
C_{1}\left(\boldsymbol{s}_{t},\boldsymbol{a}_{t}\right)=\boldsymbol{s}_{t}^{\mathrm{T}}Q_{1}\boldsymbol{s}_{t}+\boldsymbol{a}_{t}^{\mathrm{T}}R_{1}\boldsymbol{a}_{t},
\]
where $\left\{ Q_{0},R_{0},Q_{1},R_{1}\right\} $ are semi-positive
definite matrices, the state $\boldsymbol{s}_{t+1}=\mathbf{X}\boldsymbol{s}_{t}+\mathbf{Y}\boldsymbol{a}_{t}+\epsilon_{t}$
with transition matrices $\mathbf{X}\in\mathbb{R}^{n_{\boldsymbol{s}}\times n_{\boldsymbol{a}}}$
and $\mathbf{Y}\in\mathbb{R}^{n_{\boldsymbol{s}}\times n_{\boldsymbol{a}}}$,
and $\left\{ \epsilon_{t}\right\} $ is the process noise.
\begin{rem}
\textit{(Infinite-horizon average reward/cost criterion:)} There are
two reward/cost criteria are studied in RL algorithms, i.e., the infinite-horizon
average reward/cost criterion and the $\gamma$-discounted reward/cost
criterion. We adopt the former mainly because it is more challenging
in theoretical analysis. Convergence analysis techniques for the discounted
reward setting, such as backward induction and $\gamma$-contraction,
cannot be applied to the infinite-horizon average reward/cost setting
\cite{average1}. In contrast, the extension from the former to the
latter is generally considered trivial \cite{reviewer1_4_3}, both
in algorithmic design and theoretical analysis.
\end{rem}

\subsection{Neural Network Parameterization\label{subsec:Neural-Network-Parametrization}}

Following the same standard setup implemented in line with recent
works \cite{Cao2019a,Allen2019b,Cao2019b}, we consider such an $L$-hidden-layer
neural network:
\begin{equation}
f_{m}\left(\boldsymbol{\alpha};\boldsymbol{x}\right)=\sqrt{m}\mathbf{W}_{L}\sigma\left(\mathbf{W}_{L-1}\cdots\sigma\left(\mathbf{W}_{1}\boldsymbol{\phi}\bigl(\boldsymbol{x}\bigr)\right)\cdots\right),\label{eq:DNN function}
\end{equation}
where the subscript $m$ is the width of the neural network, $\sigma\left(\cdot\right)$
is the entry-wise activation function, $\boldsymbol{\phi}\left(\cdot\right)$
is a feature mapping of the input data $\boldsymbol{x}\in\mathbb{R}^{d}$,
and the input data is usually normalized, i.e., $\left\Vert \boldsymbol{x}\right\Vert _{2}\leq1$.
Moreover, $\mathbf{W}_{1}\in\mathbb{R}^{m\times d_{\mathrm{in}}}$,
$\mathbf{W}_{L}\in\mathbb{R}^{d_{\mathrm{out}}\times m}$ and $\mathbf{W}_{l}\in\mathbb{R}^{m\times m}$
for $l=2,\ldots,L-1$ are parameter matrices, and $\boldsymbol{\alpha}=\left(\mathrm{vec}\left(\mathbf{W}_{1}\right)^{\mathrm{\intercal}},\ldots,\mathrm{vec}\left(\mathbf{W}_{L}\right)^{\mathrm{\intercal}}\right)$
is the concatenation of the vectorization of all the parameter matrices.
It is easy to verify that $\left\Vert \boldsymbol{\alpha}-\boldsymbol{\alpha}'\right\Vert _{2}^{2}=\sum_{l=1}^{L}\bigl\Vert\mathbf{W}_{l}-\mathbf{W}_{l}^{'}\bigr\Vert_{F}^{2}$.
For simplicity, we assume that different layers in a network have
the same width $m$. Remark that our results can be easily generalized
to many other activation functions and the setting that the widths
of each layer are not equal.

In this paper, we parameterize both the policy and Q-functions with
the DNNs defined above. Specifically, in the actor module, we employ
the commonly used Gaussian policy \cite[Chapter 13.7]{Gaussianpolicy}
with mean $\boldsymbol{\mu}$ and diagonal elements of the covariance
matrix $\mathbf{\Sigma}$ parameterized by $f_{m_{\boldsymbol{\mu}}}\bigl(\boldsymbol{\theta}_{\boldsymbol{\mu}};\boldsymbol{s}\bigr)=\mathbb{R}^{n_{\boldsymbol{a}}}$
and $f_{m_{\boldsymbol{\sigma}}}\bigl(\boldsymbol{\theta}_{\boldsymbol{\sigma}};\boldsymbol{s}\bigr)\subseteq\mathbb{R}^{n_{\boldsymbol{a}}}$,
respectively, and keep the non-diagonal elements of $\mathbf{\Sigma}$
as $0$. That is,
\begin{equation}
\pi_{\boldsymbol{\theta}}\left(\boldsymbol{a}\mid\boldsymbol{s}\right)\propto\left|\mathbf{\Sigma}\right|^{\text{\ensuremath{-\frac{1}{2}}}}\mathrm{exp}\Bigl(-\frac{1}{2}\bigl(\boldsymbol{\mu}-\boldsymbol{a}\bigr)^{\intercal}\mathbf{\Sigma}^{-1}\bigl(\boldsymbol{\mu}-\boldsymbol{a}\bigr)\Bigr),\label{eq:pi}
\end{equation}
where the policy parameter $\boldsymbol{\theta}=\bigl[\boldsymbol{\theta}_{\boldsymbol{\mu}},\boldsymbol{\theta}_{\boldsymbol{\sigma}}\bigr]\in\mathbf{\Theta}$.
In the critic module, we adopt dual critic DNNs $f_{m_{Q}}\left(\boldsymbol{\omega}^{i};\boldsymbol{s},\boldsymbol{a}\right)\subseteq\mathbb{R}$
and $f_{m_{Q}}\left(\bar{\boldsymbol{\omega}}^{i};\boldsymbol{s},\boldsymbol{a}\right)\subseteq\mathbb{R}$
to approximate the $i$-th Q function. Please note that different
Q functions are parameterized by the same DNNs, but with different
parameters. For simplification, we abbreviate $f_{m_{Q}}\left(\boldsymbol{\omega};\boldsymbol{s},\boldsymbol{a}\right)$
to $f\left(\boldsymbol{\omega}\right)$ throughout this paper when
no confusion arises.

\subsection{Important Assumptions on the Problem Structure}

\noindent \newtheorem{assumption}{Assumption}
\begin{assumption}(Assumptions on the Problem Structure)\rm\textit{(Assumptions on the Problem Structure:)}\\
1) There are constants $\lambda>0$ and $\rho\in\left(0,1\right)$
satisfying
\begin{equation}
\mathrm{sup}_{\boldsymbol{s}\in\mathcal{S}}d_{TV}\left(\mathbf{P}\left(\boldsymbol{s}_{t}\mid\boldsymbol{s}_{0}=\boldsymbol{s}\right),\mathbf{P_{\pi_{\boldsymbol{\theta}}}}\right)\leq\lambda\rho^{t},
\end{equation}
for all $t=0,1,\cdots$, where $\mathbf{P_{\pi_{\boldsymbol{\theta}}}}$
is the stationary state distribution under policy $\pi_{\boldsymbol{\theta}}$
and $d_{TV}\left(\mu,v\right)=\int_{\boldsymbol{s}\in\mathcal{S}}\left|\mu\left(\mathrm{d}\boldsymbol{s}\right)-v\left(\mathrm{d}\boldsymbol{s}\right)\right|$
denotes the total-variation distance between the probability measures
$\mu$ and $v$.\\
2) State space $\mathcal{S}\subseteq\mathbb{R}^{n_{\boldsymbol{s}}}$
and action space $\mathcal{A}\subseteq\mathbb{R}^{n_{\boldsymbol{a}}}$
are both compact. The costs/rewards $C_{i},\forall i$, are bounded.\\
3) The gradients of $J_{i}\left(\boldsymbol{\theta}\right),\forall i$,
are uniformly bounded and follow Lipschitz continuity over the parameter
$\boldsymbol{\theta}\in\mathbf{\Theta}$.\\
4) The DNNs' parameter spaces $\mathbf{\Theta}\subseteq\mathbb{R}^{n_{\boldsymbol{\theta}}}$
and $\Omega_{i}\subseteq\mathbb{R}^{n_{\boldsymbol{\omega}}},\forall i$
are compact and convex, and the outputs of DNNs are bounded.\\
5) The policy $\pi_{\boldsymbol{\theta}}$ follows Lipschitz continuity
over the parameter $\boldsymbol{\theta}\in\mathbf{\Theta}$.\end{assumption}

Assumption 1-1) controls the bias caused by the Markovian noise in
the observations by assuming the uniform ergodicity of the Markov
chain generated by $\pi_{\boldsymbol{\theta}}$, which is a standard
requirement in the literature, see e.g., \cite{DQlearning}, \cite{linnerAC}
and \cite{SCAOPO}. Assumption 1-2) considers a general scenario in
which the state and action spaces can be continuous. Assumption 1-3)
holds based on the condition that the gradient of DNNs with respect
to parameters are bounded and satisfy Lipschitz continuity. Remark
this condition can be satisfied by a large category of DNNs with Lipschitz-smooth
activation functions and over-parameterized DNNs using the rectified
linear unit (ReLU) function \cite{Allen2019b}. Assumption 1-4) is
trivial in CRL problems. Assumption 1-5) can be easily satisfied as
long as there is no gradient explosion during the training process.
Some techniques such as proper initialization and gradient clipping
can help avoid gradient explosion \cite{gradient0clipping,gradient1clipping,initialization}.

\section{\label{sec:Algorithmic-Framework}SLDAC Algorithmic Framework}

\begin{figure}[t]
\noindent \centering{}\includegraphics[width=8cm,height=5cm]{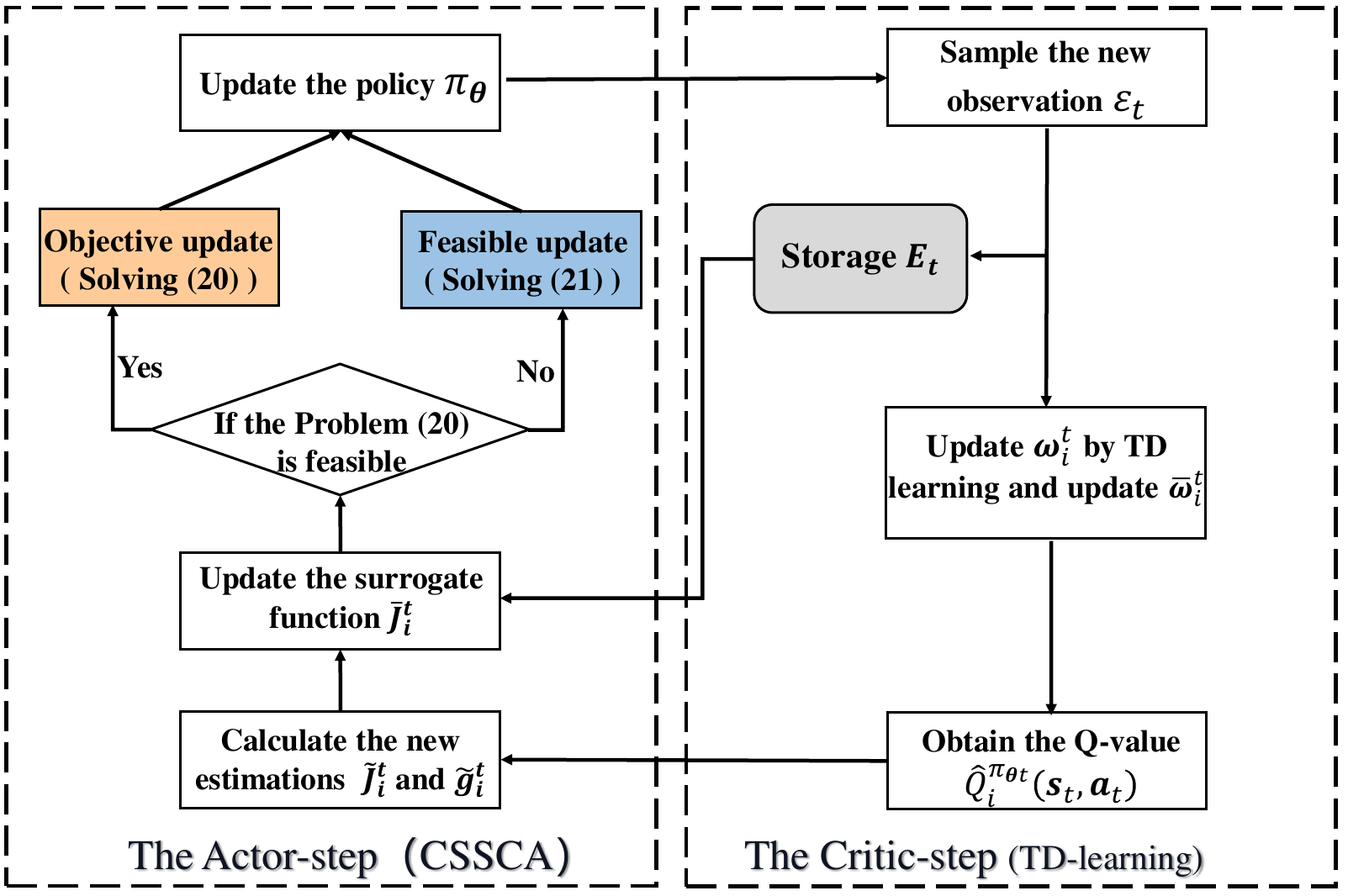}\caption{\textcolor{blue}{\label{fig:diagram of the algorithm}}The algorithmic
framework of the proposed SLDAC}
\end{figure}
 As illustrated in Fig. \ref{fig:diagram of the algorithm}, the proposed
SLDAC algorithm performs iterations between the critic module and
the actor module. At the $t$-th iteration, the new observation $\varepsilon_{t}$
is used to update the critic module, and old observations in $E_{t}$
are allowed to be reused to update the actor module. In the following,
we elaborate on the sampling and storage, as well as the design of
the actor and critic modules in detail. The overall SLDAC algorithm
is summarized in Algorithm \ref{alg:the-Single-Loop-Deep} and for
readers' convenience, the meanings of some important symbols hereafter
are listed in Table \ref{tab:Meanings-of-symbols-1}.

\subsection{Sampling and Storage}

Once given the current policy $\pi_{\boldsymbol{\theta}_{t}}$, the
agent at state $\boldsymbol{s}_{t}$ can choose an action $\boldsymbol{a}_{t}$
according to the policy $\pi_{\boldsymbol{\theta}_{t}}$ and transitions
to the next new state $\boldsymbol{s}_{t+1}$ with the environment
feeding back a set of cost (reward) function values $\bigl\{ C_{i}(\boldsymbol{s}_{t},\boldsymbol{a}_{t})\bigr\}_{i=0,\ldots,I}$.
Then, we denote the tuple $\varepsilon_{t}=\bigl\{\boldsymbol{s}_{t},\boldsymbol{a}_{t},\bigl\{ C_{i}(\boldsymbol{s}_{t},\boldsymbol{a}_{t})\bigr\}_{i=0,\ldots,I},\boldsymbol{s}_{t+1}\bigr\}$
as a new observation at the $t$-th iteration and store the latest
$T_{t}$ observations in the storage $E_{t}$, i.e., $E_{t}=\bigl\{\varepsilon_{t-T_{t}+1},\varepsilon_{t-T_{t}+2},\ldots,\varepsilon_{t}\bigr\}$.
In practice, a minibatch of $B$ new observations can be obtained
by performing this interaction repeatedly under $\pi_{\boldsymbol{\theta}_{t}}$,
to perform a few finite $q<B$ critic updates at each iteration, where
$B/q$ new observations are used in each critic update. The batch
size $B$ and the number of inner iterations $q$ can be properly
chosen to achieve a flexible tradeoff between better estimation accuracy
for the Q values/policy gradient and the interaction cost with the
environment as well as the computational complexity. In the algorithm
design and convergence analyses of this paper, we only focus on the
most challenging case when $B=q=1$ for clarity. However, the scheme
and theoretical result also hold for arbitrary choice of $B\geq q\geq1$. 

\subsection{The Critic Module}

We first denote $C_{0}^{\text{'}}(\boldsymbol{s}_{t},\boldsymbol{a}_{t})=C_{0}(\boldsymbol{s}_{t},\boldsymbol{a}_{t})$
and $C_{i}^{\text{'}}(\boldsymbol{s}_{t},\boldsymbol{a}_{t})=C_{i}(\boldsymbol{s}_{t},\boldsymbol{a}_{t})-c_{i},i=1,\ldots,I$.
For any $\pi_{\boldsymbol{\theta}_{t}}$, there is a set of Q functions
$\bigl\{ Q_{i}^{\pi_{\boldsymbol{\theta}_{t}}}\bigr\}_{i=0,,\ldots I}$
defined as
\begin{align}
Q_{i}^{\pi_{\boldsymbol{\theta}_{t}}}\left(\boldsymbol{s},\boldsymbol{a}\right)= & \mathbb{E}_{p_{t}}\Bigl[\sum_{l=0}^{\infty}\Bigl(C_{i}^{\text{'}}(\boldsymbol{s}_{l},\boldsymbol{a}_{l})-J_{i}\left(\boldsymbol{\theta}_{t}\right)\Bigr)\nonumber \\
 & \bigl|\boldsymbol{s}_{0}=\boldsymbol{s},\boldsymbol{a}_{0}=\boldsymbol{a}\Bigr],\forall i,\boldsymbol{s},\boldsymbol{a},
\end{align}
where we denote the distribution $p_{\pi_{\boldsymbol{\theta}_{t}}}$
by $p_{t}$ for short. Since it is unrealistic to obtain $J_{i}\left(\boldsymbol{\theta}_{t}\right)$
online, we redefine a set of surrogate Q functions $\bigl\{\hat{Q}_{i}^{\pi_{\boldsymbol{\theta}_{t}}}\bigr\}_{i=0,,\ldots I}$
as
\begin{align}
\hat{Q}_{i}^{\pi_{\boldsymbol{\theta}_{t}}}\left(\boldsymbol{s},\boldsymbol{a}\right)= & \mathbb{E}_{p_{t}}\Bigl[\sum_{l=0}^{\infty}\Bigl(C_{i}^{\text{'}}(\boldsymbol{s}_{l},\boldsymbol{a}_{l})-\hat{J}_{i}^{t}\Bigr)\nonumber \\
 & \bigl|\boldsymbol{s}_{0}=\boldsymbol{s},\boldsymbol{a}_{0}=\boldsymbol{a}\Bigr],\forall i,\boldsymbol{s},\boldsymbol{a},\label{eq:definition of Qhat}
\end{align}
where $\hat{J}_{i}^{t}$ is the estimate of $J_{i}\left(\boldsymbol{\theta}_{t}\right)$
that will be given in (\ref{eq:J-average}) and be proved that $\lim_{t\rightarrow\infty}\bigl|\hat{J}_{i}^{t}-J_{i}\bigl(\boldsymbol{\theta}_{t}\bigr)\bigr|=0$.
To approximate $\bigl\{\hat{Q}_{i}^{\pi_{\boldsymbol{\theta}_{t}}}\bigr\}_{i=0,,\ldots I}$,
we adopt two sets of critic DNNs and update them as follows.

The first set of DNNs $\bigl\{ f\bigl(\boldsymbol{\omega}^{i}\bigr)\bigr\}_{i=0,\ldots,I}$
is updated to minimize the mean-squared projected Bellman error (MSBE)
\cite{DQlearning}:
\begin{equation}
\underset{\boldsymbol{\omega}_{t}^{i}}{\mathrm{min}}\ \mathbb{E}_{\sigma_{\pi_{\boldsymbol{\theta}_{t}}}}\Bigl[\Bigl(f\left(\boldsymbol{\omega}_{t}^{i};\boldsymbol{s}_{t},\boldsymbol{a}_{t}\right)-\mathcal{T}_{t}\ f\left(\boldsymbol{\omega}_{t}^{i};\boldsymbol{s}_{t},\boldsymbol{a}_{t}\right)\Bigr)^{2}\Bigr],\forall i,\label{eq:MSBE}
\end{equation}
where $\sigma_{\pi_{\boldsymbol{\theta}_{t}}}\left(\boldsymbol{s}_{t},\boldsymbol{a}_{t}\right)=\pi_{\boldsymbol{\theta}_{t}}\left(\boldsymbol{a}_{t}\mid\boldsymbol{s}_{t}\right)\cdot\mathbf{P_{\pi}}\left(\boldsymbol{s}_{t}\right)$
is the stationary state-action distribution, and we abbreviate it
as $\sigma_{t}$. In addition, the Bellman operator $\mathcal{T}_{t}$
is defined as
\begin{align}
\mathcal{T}_{t}\ f\left(\boldsymbol{\omega}_{t}^{i};\boldsymbol{s}_{t},\boldsymbol{a}_{t}\right)= & \mathbb{E}_{p_{t}}\Bigl[f\left(\boldsymbol{\omega}_{t}^{i};\boldsymbol{s}_{t+1},\boldsymbol{a}_{t+1}'\right)\Bigr]\\
 & +C_{i}^{\text{'}}(\boldsymbol{s}_{t},\boldsymbol{a}_{t})-\hat{J}_{i}^{t},\forall i,\nonumber 
\end{align}
where $\boldsymbol{a}_{t+1}'$ is the action chosen by current policy
$\pi_{\boldsymbol{\theta}_{t}}$ at the next state $\boldsymbol{s}_{t+1}$.
To solve Problem (\ref{eq:MSBE}), we first define a neighborhood
of the randomly initialized parameter $\boldsymbol{\omega}_{0}^{i},\forall i$
as
\begin{align}
\mathbb{B}\left(\boldsymbol{\omega}_{0}^{i},R_{\boldsymbol{\omega}}\right)= & \Bigl\{\boldsymbol{\omega}^{i}=\bigl(\mathrm{vec}\bigl(\mathbf{W}_{1}^{i}\bigr)^{\top},\ldots,\mathrm{vec}\bigl(\mathbf{W}_{L}^{i}\bigr)^{\top}\bigr)^{\top}:\nonumber \\
 & \bigl\Vert\mathbf{W}_{l}^{i}-\mathbf{W}_{l}^{i,(0)}\bigr\Vert_{F}\leq R_{\boldsymbol{\omega}},l=1,\ldots,L\Bigr\},\label{eq:constraint set}
\end{align}
where $\mathbf{W}_{l}^{i,(0)}$ is the parameter of the $l$-th layer
corresponding to $\boldsymbol{\omega}_{0}^{i}$. Then, we update the
parameter $\boldsymbol{\omega}^{i}$ in the neighborhood $\mathbb{B}\left(\boldsymbol{\omega}_{0}^{i},R_{\boldsymbol{\omega}}\right)$
using the TD-learning method:
\begin{equation}
\boldsymbol{\omega}_{t}^{i}=\Pi_{\Omega_{i}}\bigl(\boldsymbol{\omega}_{t-1}^{i}-\eta_{t}\boldsymbol{\Delta}_{i}^{\boldsymbol{\omega}_{t-1}^{i}}\bigr),\forall i,\label{eq:TD-learning}
\end{equation}
where $\left\{ \eta_{t}\right\} $ is a decreasing sequence satisfying
Assumption 2 in section \ref{subsec:Assumptions-on-stepsizes}, $\Pi_{\Omega_{i}}$
is a projection operator that projects the parameter into the constraint
set $\Omega_{i}\triangleq\mathbb{B}\left(\boldsymbol{\omega}_{0}^{i},R_{\boldsymbol{\omega}}\right)$,
and the stochastic gradient $\boldsymbol{\Delta}_{i}^{\boldsymbol{\omega}_{t-1}^{i}},\forall i$
is defined as
\begin{align}
\boldsymbol{\Delta}_{i}^{\boldsymbol{\omega}_{t-1}^{i}}= & \Bigl(f\left(\boldsymbol{\omega}_{t-1}^{i};\boldsymbol{s}_{t},\boldsymbol{a}_{t}\right)-\bigl(C_{i}^{\text{'}}(\boldsymbol{s}_{t},\boldsymbol{a}_{t})-\hat{J}_{i}^{t-1}\label{eq:gradient term}\\
 & +f\left(\boldsymbol{\omega}_{t-1}^{i};\boldsymbol{s}_{t+1},\boldsymbol{a}_{t+1}'\right)\bigl)\Bigr)\nabla_{\boldsymbol{\omega}}f\left(\boldsymbol{\omega}_{t-1}^{i};\boldsymbol{s}_{t},\boldsymbol{a}_{t}\right).\nonumber 
\end{align}

To enhance stability, we adopt the other set of critic networks $\Bigl\{ f\left(\bar{\boldsymbol{\omega}}^{i}\right)\Bigr\}_{i=0,\ldots,I}$
and use their outputs to calculate the policy gradients in the actor
module. The parameter $\bar{\boldsymbol{\omega}}^{i}$ is updated
by the following recursive operation from the $0$-th iteration $\bar{\boldsymbol{\omega}}_{0}^{i}=\boldsymbol{\omega}_{1}^{i}$:
\begin{equation}
\bar{\boldsymbol{\omega}}_{t}^{i}=\left(1-\gamma_{t}\right)\bar{\boldsymbol{\omega}}_{t-1}^{i}+\gamma_{t}\boldsymbol{\omega}_{t}^{i},\forall i,\label{eq:w^_}
\end{equation}
where $\left\{ \gamma_{t}\right\} $ is also decreasing sequence satisfying
the Assumption 2 in section \ref{subsec:Assumptions-on-stepsizes}.
\begin{table}[th]
\centering{}\caption{\textcolor{blue}{\label{tab:Meanings-of-symbols-1}}Meanings of important
symbols}
\begin{tabular}{|c|c|}
\hline 
{\scriptsize{}Symbol} & {\scriptsize{}Meaning}\tabularnewline
\hline 
\hline 
{\scriptsize{}$\tilde{J}_{i}^{t}$/$\tilde{\boldsymbol{g}}_{i}^{t}$} & {\scriptsize{}New estimate of the function value/gradient to update
$\hat{J}_{i}^{t}$/$\hat{\boldsymbol{g}}_{i}^{t}$}\tabularnewline
\hline 
{\scriptsize{}$\hat{J}_{i}^{t}$/$\hat{\boldsymbol{g}}_{i}^{t}$} & {\scriptsize{}Estimate of the function value/gradient to construct
$\bar{J}_{i}^{t}$}\tabularnewline
\hline 
{\scriptsize{}$J_{i}\bigl(\boldsymbol{\theta}_{t}\bigr)$, $\bar{J}_{i}^{t}$} & {\scriptsize{}Original objective function and its surrogate function}\tabularnewline
\hline 
{\scriptsize{}$Q_{i}^{\pi_{\boldsymbol{\theta}_{t}}}$, $\hat{Q}_{i}^{\pi_{\boldsymbol{\theta}_{t}}}$} & {\scriptsize{}Exact Q-function and its surrogate function calculated
by $\hat{J}_{i}^{t}$}\tabularnewline
\hline 
{\scriptsize{}$\boldsymbol{\omega}_{t},\bar{\boldsymbol{\omega}}_{t}$} & {\scriptsize{}Parameters for the dual sets of critic networks}\tabularnewline
\hline 
{\scriptsize{}$f\bigl(\boldsymbol{\omega}_{t}^{i}\bigr)$, $f\bigl(\bar{\boldsymbol{\omega}}_{t}^{i}\bigr)$} & {\scriptsize{}Dual sets of critic networks to approximate $\hat{Q}_{i}^{\pi_{\boldsymbol{\theta}_{t}}}$}\tabularnewline
\hline 
{\scriptsize{}$\boldsymbol{m},\bar{\boldsymbol{m}}$} & {\scriptsize{}Auxiliary parameters for $\boldsymbol{\omega}_{t}$
and $\bar{\boldsymbol{\omega}}_{t}$}\tabularnewline
\hline 
{\scriptsize{}$\hat{f}\bigl(\bar{\boldsymbol{\omega}}_{t}^{i}\bigr)$} & {\scriptsize{}Local linearization function of $f\bigl(\bar{\boldsymbol{\omega}}_{t}^{i}\bigr)$}\tabularnewline
\hline 
{\scriptsize{}$\varepsilon_{t}$, $E_{t}$} & {\scriptsize{}New observation and its storage under $\pi_{\boldsymbol{\theta}_{t}}$}\tabularnewline
\hline 
{\scriptsize{}$\tilde{\varepsilon}_{t}$, $\tilde{E}_{t}$} & {\scriptsize{}Auxiliary observation and its storage under fixed $\pi_{\boldsymbol{\theta}_{t-n_{t}}}$}\tabularnewline
\hline 
\end{tabular}\vspace{-0.5cm}
\end{table}

\subsection{The Actor Module}

The key to solving (\ref{eq:problem 1}) in the actor step is to replace
the objective/constraint functions $\bigl\{ J_{i}\left(\boldsymbol{\theta}\right),\forall i\bigr\}$
by some convex surrogate functions $\bigl\{\bar{J}_{i}^{t}\left(\boldsymbol{\theta}\right),\forall i\bigr\}$
constructed by the estimated function values $\bigl\{\hat{\boldsymbol{J}}_{i}^{t},\forall i\bigr\}$
and the estimated policy gradients $\bigl\{\hat{\boldsymbol{g}}_{i}^{t},\forall i\bigr\}$.
Then, the original problem is addressed by solving a sequence of convex
optimization problems.

The surrogate function $\bar{J}_{i}^{t}\left(\boldsymbol{\theta}\right)$
can be seen as a convex approximation of $J_{i}\left(\boldsymbol{\theta}\right)$
based on the $t$-th iterate $\boldsymbol{\theta}_{t}$, which is
formulated as:
\begin{equation}
\bar{J}_{i}^{t}\left(\boldsymbol{\theta}\right)=\hat{J}_{i}^{t}+\left(\hat{\boldsymbol{g}}_{i}^{t}\right)^{\textrm{\ensuremath{\intercal}}}\left(\boldsymbol{\theta}-\boldsymbol{\theta}_{t}\right)+\zeta_{i}\left\Vert \boldsymbol{\theta}-\boldsymbol{\theta}_{t}\right\Vert _{2}^{2},\forall i,\label{eq:surrogate functions}
\end{equation}
where $\zeta_{i}$ is a positive constant, $\hat{J}_{i}^{t}\in\mathbb{R}$
is the estimate of $J_{i}\left(\boldsymbol{\theta}\right)$, and $\hat{\boldsymbol{g}}_{i}^{t}\in\mathbb{R}^{n_{\boldsymbol{\theta}_{i}}}$
is the estimate of gradient $\nabla J_{i}\left(\boldsymbol{\theta}\right)$
at the $t$-th iteration, which are updated by
\begin{equation}
\hat{J}_{i}^{t}=\left(1-\alpha_{t}\right)\hat{J}_{i}^{t-1}+\alpha_{t}\tilde{J}_{i}^{t},\forall i,\label{eq:J-average}
\end{equation}
\begin{equation}
\hat{\boldsymbol{g}}_{i}^{t}=\left(1-\alpha_{t}\right)\hat{\boldsymbol{g}}_{i}^{t-1}+\alpha_{t}\tilde{\boldsymbol{g}}_{i}^{t},\forall i,\label{eq:g-average}
\end{equation}
where the step size $\left\{ \alpha_{t}\right\} $ is a decreasing
sequence satisfying the Assumption 2 in Section \ref{sec:CONVERGENCE-ANALYSIS},
$\tilde{J}_{i}^{t}$ and $\tilde{\boldsymbol{g}}_{i}^{t}$ are the
realizations of function value and its gradient whose specific forms
are given below.

To reduce the estimation variance and use the observations more efficiently,
we propose an off-policy estimation strategy in which $\tilde{J}_{i}^{t}$
and $\tilde{\boldsymbol{g}}_{i}^{t}$ can be obtained by reusing old
observations in $E_{t}$. By the sample average method, we can first
obtain the new estimation of function value at the $t$-th iteration
$\tilde{J}_{i}^{t}$:
\begin{equation}
\tilde{J}_{i}^{t}=\frac{1}{T_{t}}\stackrel[l=1]{T_{t}}{\sum}C_{i}^{\text{'}}(\boldsymbol{s}_{t-T_{t}+l},\boldsymbol{a}_{t-T_{t}+l}),\forall i.\label{eq:J-new}
\end{equation}
Then according to the policy gradient theorem \cite{RLintroduction},
we have
\begin{equation}
\nabla J_{i}\left(\boldsymbol{\theta}\right)=\mathbb{E}_{\sigma_{\pi_{\boldsymbol{\theta}}}}\left[Q_{i}^{\pi_{\boldsymbol{\theta}}}\left(\boldsymbol{s},\boldsymbol{a}\right)\nabla_{\boldsymbol{\theta}}\textrm{log}\pi_{\boldsymbol{\theta}}\left(\boldsymbol{a}\mid\boldsymbol{s}\right)\right],\forall i.\label{eq:real policy gradient}
\end{equation}
We adopt the idea of the sample average and give the estimate of the
gradient at the $t$-th iteration as
\begin{align}
\tilde{\boldsymbol{g}}_{i}^{t}= & \frac{1}{T_{t}}\stackrel[l=1]{T_{t}}{\sum}f\left(\bar{\boldsymbol{\omega}}_{t}^{i};\boldsymbol{s}_{t-T_{t}+l},\boldsymbol{a}_{t-T_{t}+l}\right)\label{eq:g-tilde}\\
 & \cdot\nabla_{\boldsymbol{\theta}_{t}}\textrm{log}\pi_{\boldsymbol{\theta}_{t}}\left(\boldsymbol{a}_{t-T_{t}+l}\mid\boldsymbol{s}_{t-T_{t}+l}\right),\forall i.\nonumber 
\end{align}

Based on the surrogate functions $\left\{ \bar{J}_{i}^{t}\left(\boldsymbol{\theta}\right)\right\} _{i=0,\ldots,I}$,
the optimal solution $\bar{\boldsymbol{\theta}}_{t}$ of the following
problem is solved:
\begin{align}
\bar{\boldsymbol{\theta}}_{t}=\underset{\boldsymbol{\theta}\in\mathbf{\Theta}}{\textrm{argmin}}\  & \bar{J}_{0}^{t}\left(\boldsymbol{\theta}\right)\label{eq:objective update}\\
\textrm{s.t.}\  & \bar{J}_{i}^{t}\left(\boldsymbol{\theta}\right)\leq0,i=1,\ldots,I.\nonumber 
\end{align}
If problem (\ref{eq:objective update}) turns out to be infeasible,
the optimal solution $\bar{\boldsymbol{\theta}}_{t}$ of the following
convex problem is solved:
\begin{align}
\bar{\boldsymbol{\theta}}_{t}=\underset{\boldsymbol{\theta}\in\mathbf{\Theta},y}{\textrm{argmin}}\  & y\label{eq:feasible update}\\
\textrm{s.t.}\  & \bar{J}_{i}^{t}\left(\boldsymbol{\theta}\right)\leq y,i=1,\cdots,I.\nonumber 
\end{align}
Please note that the surrogate problems (\ref{eq:objective update})
and (\ref{eq:feasible update}) both belong to the convex quadratic
problem with a closed-form solution, which can be easily solved by
standard convex optimization algorithms, e.g. Lagrange-dual methods.
Then, once given $\bar{\boldsymbol{\theta}}_{t}$, $\boldsymbol{\theta}_{t+1}$
is updated according to
\begin{equation}
\boldsymbol{\theta}_{t+1}=(1-\beta_{t})\boldsymbol{\theta}_{t}+\beta_{t}\bar{\boldsymbol{\theta}}_{t}.\label{eq:theta update}
\end{equation}
where the step size $\left\{ \beta_{t}\right\} $ is a decreasing
sequence satisfying Assumption 2 in Section \ref{sec:CONVERGENCE-ANALYSIS}.
\begin{algorithm}
\caption{\label{alg:the-Single-Loop-Deep}Single-Loop Deep Actor-Critic Algorithm}

\textbf{Input:} The decreasing sequences $\left\{ \alpha_{t}\right\} $,
$\left\{ \beta_{t}\right\} $, $\left\{ \eta_{t}\right\} $, and $\left\{ \gamma_{t}\right\} $,
randomly generate the initial entries of $\boldsymbol{\omega}_{0}^{i}$
and $\boldsymbol{\theta}_{0}$ from $\mathcal{N}\left(0,1/m^{2}\right)$.

\textbf{for} $t=0,1,\cdots$ \textbf{do}

\phantom{} \phantom{}\textbf{ }Sample the new observation $\varepsilon_{t}$
and update the storage $E_{t}$.

\phantom{} \phantom{}\textbf{ Critic Step:}

\phantom{} \phantom{}\textbf{ }\phantom{} \phantom{} Update $\boldsymbol{\omega}_{t}^{i}$
and $\bar{\boldsymbol{\omega}}_{t}^{i}$ by (\ref{eq:TD-learning})
and (\ref{eq:w^_}), respectively.

\phantom{} \phantom{}\textbf{ Actor Step:}

\phantom{} \phantom{}\textbf{ }\phantom{} \phantom{} Calculate
$\hat{J}_{i}^{t}$ according to (\ref{eq:J-new}) and (\ref{eq:J-average}).

\phantom{} \phantom{}\textbf{ }\phantom{} \phantom{} Estimate
gradient $\hat{\boldsymbol{g}}_{i}^{t}$ according to (\ref{eq:g-tilde})
and (\ref{eq:g-average}).

\phantom{} \phantom{}\textbf{ }\phantom{} \phantom{} Update the
surrogate function $\left\{ \bar{J}_{i}^{t}\left(\boldsymbol{\theta}\right)\right\} _{i=0,\ldots,I}$via
(\ref{eq:surrogate functions}).

\phantom{} \phantom{}\textbf{ }\phantom{} \phantom{} \textbf{if
}Problem (\ref{eq:objective update}) is feasible:

\phantom{} \phantom{}\textbf{ }\phantom{} \phantom{} \phantom{}
\phantom{} Solve (\ref{eq:objective update}) to obtain $\bar{\boldsymbol{\theta}}_{t}$.

\phantom{} \phantom{}\textbf{ }\phantom{} \phantom{} \textbf{else}

\phantom{} \phantom{}\textbf{ }\phantom{} \phantom{} \phantom{}
\phantom{} Solve (\ref{eq:feasible update}) to obtain $\bar{\boldsymbol{\theta}}_{t}$.

\phantom{} \phantom{}\textbf{ }\phantom{} \phantom{} \textbf{end
if}

\phantom{} \phantom{}\textbf{ }\phantom{} \phantom{} Update policy
parameters $\boldsymbol{\theta}_{t+1}$ according to (\ref{eq:theta update}).

\textbf{end for}
\end{algorithm}

\subsection{Key Differences from Closely Related Works}

As we mentioned in section \ref{subsec:Methods-for-CMDP}, several
advanced CRL algorithms have been designed, e.g. TRPO-Lag and PPO-Lag
\cite{PPOLagTRPOLag}, CPO \cite{CPO}, and our previous work \cite{SCAOPO}.
However, there are still key differences in algorithm design between
this paper and these closely related works, as illustrated in Table
\ref{tab:Key-differences}.

TRPO-Lag, PPO-Lag, and CPO are all advanced DAC algorithms for CMDPs.
However, they do not have a special design for step sizes as Assumption
2 in this paper and therefore belong to the two-loop AC, that is,
$B\rightarrow\infty$ and $q\rightarrow\infty$ (or $B$ and $q$
are sufficiently large in practice) are required to avoid error accumulation,
which leads to high interaction cost. Moreover, they adopt on-policy
sampling and their approaches to policy optimization make them only
suitable for simple convex constraints. Both the SCAOPO in \cite{SCAOPO}
and the SLDAC in this paper allow the observation reuse to improve
the observation efficiency, adopt the CSSCA to better handle the non-convexity
of objective and constraints and guarantee the final convergence to
a KKT point (up to an error of $\epsilon_{m_{Q}}$ with $\underset{m_{Q}\rightarrow\infty}{\mathrm{lim}}\epsilon_{m_{Q}}=0$).
However, the SCAOPO is based on the actor-only framework, which simply
adopts the MC method to estimate Q values. Instead of the MC method,
we use dual DNNs to approximate Q functions, which reduces variance
and tends to find more accurate estimates. These innovative designs
of SLDAC help to achieve superior performance with much lower interaction
cost, although it makes convergence analysis more challenging. 
\begin{table}
\centering{}\caption{\textcolor{blue}{\label{tab:Key-differences}}Key differences from
closely related works}
{\scriptsize{}}%
\begin{tabular}{|>{\centering}m{2.3cm}|>{\centering}m{1.6cm}|>{\centering}m{1.3cm}|>{\centering}m{1.7cm}|}
\hline 
{\scriptsize{}Algorithm for CMDPs} & {\scriptsize{}Framework} & {\scriptsize{}Observation reuse} & {\scriptsize{}Convergence to a KKT point}\tabularnewline
\hline 
\hline 
{\scriptsize{}DAC in \cite{PPOLagTRPOLag}, \cite{CPO}} & {\scriptsize{}Two-loop AC} & {\scriptsize{}\texttimes{}} & {\scriptsize{}\texttimes{}}\tabularnewline
\hline 
{\scriptsize{}SCAOPO in \cite{SCAOPO}} & {\scriptsize{}Actor-only} & {\scriptsize{}\textsurd{}} & {\scriptsize{}\textsurd{}}\tabularnewline
\hline 
{\scriptsize{}SLDAC in this paper} & {\scriptsize{}Single-loop AC} & {\scriptsize{}\textsurd{}} & {\scriptsize{}\textsurd{}}\tabularnewline
\hline 
\end{tabular}
\end{table}

\section{Convergence Analysis\label{sec:CONVERGENCE-ANALYSIS}}

In the following, we first present the key assumptions on step sizes
and then analyze the convergence rate of the critic module. Based
on this theoretical result and assumptions above, we further show
that the surrogate functions satisfy asymptotic consistency. Finally,
we prove that the proposed Algorithm 1 converges to a KKT point (up
to an error of $\epsilon_{m_{Q}}$ with $\underset{m_{Q}\rightarrow\infty}{\mathrm{lim}}\epsilon_{m_{Q}}=0$)
of the original Problem (1).

\subsection{Key Assumptions on Step Sizes\label{subsec:Assumptions-on-stepsizes}}

To state the convergence results, we need to lay down some assumptions
on the sequence of step sizes:

\begin{assumption}\rm\textit{ (Assumptions on step size:)}\\
The step-sizes $\left\{ \alpha_{t}\right\} $, $\left\{ \beta_{t}\right\} $,
$\left\{ \eta_{t}\right\} $ and $\left\{ \gamma_{t}\right\} $ are
deterministic and non-increasing, and satisfy:\\
1) $\alpha_{t}\rightarrow0$, $\frac{1}{\alpha_{t}}\leq O\left(t^{\kappa}\right)$
for some $\kappa\in\left(0,1\right)$, $\sum_{t}\alpha_{t}t^{-1}<\infty$,
$\sum_{t}\left(\alpha_{t}\right)^{2}<\infty$.\\
2) $\beta_{t}\rightarrow0$, $\sum_{t}\beta_{t}=\infty$, $\sum_{t}\left(\beta_{t}\right)^{2}<\infty$,
$\mathrm{lim}_{t\rightarrow\infty}\beta_{t}\alpha_{t}^{-1}=0$,\\
$\sum_{t}\alpha_{t}\beta_{t}\mathrm{log}t<\infty$.\\
3) $\eta_{t}\rightarrow0$, $\gamma_{t}\rightarrow0$, $\sum_{t}\eta_{t}=\infty$,
$\sum_{t}\gamma_{t}=\infty$.\\
4) $\sum_{t}\alpha_{t}\bigl(1-\gamma_{t}\bigr)^{t^{0.215}}\gamma_{n_{t}}^{-1/2}<\infty$,
$\sum_{t}\ensuremath{\alpha_{t}\gamma_{n_{t}}^{1/2}\eta_{n_{t}}^{-1/2}<\infty}$,\\
$\sum_{t}m_{Q}\alpha_{t}\gamma_{n_{t}}^{1/2}\eta_{n_{t}}^{1/2}t^{0.215}<\infty$,
$\sum_{t}m_{Q}\alpha_{t}\eta_{n_{t}}t^{-0.57}<\infty$, \\
and $\sum_{t}m_{Q}\alpha_{t}\eta_{n_{t}}\beta_{n_{t}}t^{0.86}<\infty$,
where $n_{t}=t-t^{0.43}$.

\end{assumption} Please note that Assumptions 2.1)-2.3) are common
in the works \cite{CSSCA}, \cite{SCAOPO}, and \cite{reviewer1_4_3}.
On the other hand, Assumption 2.4) is newly added to ensure that the
critic module converges fast enough for asymptotic consistency in
section \ref{subsec:Asymptotic-Consistency} to be satisfied. Although
Assumptions 2.4) seems complicated, it is crucial. To make the step
size assumptions as intuitive as possible, we set the step size $\alpha_{t}=O\bigl(m_{Q}^{-1/2}t^{-\kappa_{1}}\bigr)$,
$\beta_{t}=O\bigl(m_{Q}^{-1/2}t^{-\kappa_{2}}\bigr)$, $\eta_{t}=O\bigl(m_{Q}^{-1/2}t^{-\kappa_{3}}\bigr)$,
and $\gamma_{t}=O\bigl(t^{-\kappa_{4}}\bigr)$, and then the Assumption
2 can be satisfied when $\kappa_{i},i=1,\ldots,4$ lies in the following
region
\[
\begin{cases}
1>2\kappa_{2}-1>\kappa_{1}>0.43>\kappa_{4}>0,\\
\mathrm{min}\left\{ 0.5\kappa_{1}+0.5\kappa_{2}+0.5\kappa_{3}-0.5,\kappa_{1}+\kappa_{3}\right\} >0.43,\\
\kappa_{1}+0.5\kappa_{4}-0.5\kappa_{3}>1,\\
\kappa_{1}+0.5\kappa_{3}+0.5\kappa_{4}>1.215.
\end{cases}
\]
In practice, we can choose $\kappa_{1},\kappa_{2},\kappa_{3},\kappa_{4}$
in the regions $\bigl(0.5,1\bigr)$, $\bigl(0.5+0.5\kappa_{1},1\bigr)$,
$\bigl(\mathrm{max}\left\{ 0.43-\kappa_{1},1.92-\kappa_{1}-\kappa_{2}\right\} ,1\bigr)$,
and $\bigl(\mathrm{max}\left\{ 2+\kappa_{3}-2\kappa_{1},2.43-2\kappa_{1}-\kappa_{3}\right\} ,0.43\bigr)$
in turn to ensure this set of inequalities satisfied, e.g. $\kappa_{1}=0.9,\kappa_{2}=0.96,\kappa_{3}=0.21$,
and $\kappa_{4}=0.42$ is a point in this region.
\begin{rem}
\textit{(Suggestion for step size selection in practice)} The above
step size region is chosen to facilitate rigorous convergence proof,
considering that it is very difficult to bound various convergence
errors due to the single-loop design and observation reuse to reduce
the interaction cost and complexity. As such, the step size conditions
in Assumption 2 are sufficient but not necessary for convergence.
The actual step size region used in practice can be wider. In fact,
we can appropriately choose larger step size parameters $\kappa_{1},\kappa_{2},\kappa_{3},\kappa_{4}$
with slower decreasing speed for the step sizes in practice, because
faster initial convergence speed is generally preferred. In the experiments,
we follow the general step-size rule in Assumption 2\textbf{ }but
slightly relax the conditions to\textbf{ }achieve a good convergence
speed. The exact values of the step sizes are given in the simulation
section for each application example.
\end{rem}

\subsection{Convergence of the Critic Module\label{subsec:Convergence-of-Critic}}

Recalling that the task of the critic module is to approximate the
surrogate Q functions $\bigl\{\hat{Q}_{i}^{\pi_{\boldsymbol{\theta}_{t}}}\bigr\}_{i=0,,\ldots I}$
using DNNs $\bigl\{ f\bigl(\bar{\boldsymbol{\omega}}^{i}\bigr)\bigr\}_{i=0,\ldots,I}$,
we derive the estimated error $\epsilon_{\mathrm{cri}}\bigl(t\bigr)\triangleq\bigl|\mathbb{E}_{p_{t}}\bigl[f\bigl(\bar{\boldsymbol{\omega}}_{t}^{i}\bigr)\bigl|\bar{\boldsymbol{\omega}}_{t-1}^{i}\bigr]-\hat{Q}_{i}^{\pi_{\boldsymbol{\theta}_{t}}}\bigr|,\forall i$
in this section. For notation simplicity, we denote $\mathbb{E}_{p_{t}}\bigl[\cdot\bigl|\bar{\boldsymbol{\omega}}_{t-1}^{i}\bigr]$
as $\mathbb{E}\bigl[\cdot\bigr]$, and the same abbreviation also
applies to other parameters and vectors below, when no confusion arises.

\subsubsection{Auxiliary Functions}

Before formally presenting the convergence result, we first define
an auxiliary function class $\mathcal{\hat{F}}$ for the intractable
convergence analysis for DNNs, as with \cite{2019b} and \cite{DQlearning}:

\noindent \newtheorem{definition}{Defination}\begin{definition}\rm
\textit{(Local Linearization Function Class) }\\
For each critic DNN function $f$, we define a function class:
\begin{align*}
\mathcal{\hat{F}}= & \bigl\{\hat{f}\bigl(\boldsymbol{\omega}\bigr)=f\bigl(\boldsymbol{\omega}_{0}\bigr)+\bigl\langle\nabla_{\boldsymbol{\omega}}f\bigl(\boldsymbol{\omega}_{0}\bigr),\boldsymbol{\omega}-\boldsymbol{\omega}_{0}\bigr\rangle:\\
 & \boldsymbol{\omega}\in\mathbb{B}\bigl(\boldsymbol{\omega}_{0},R_{\boldsymbol{\omega}}\bigr)\bigr\},
\end{align*}
where $\boldsymbol{\omega}_{0}$ is the randomly initialized parameter,
and we denote by $\bigl\langle\cdot,\cdot\bigr\rangle$ the inner
product. Then, based on $\nabla_{\boldsymbol{\omega}}f\bigl(\boldsymbol{\omega}_{0}\bigr)$,
we define a square matrix similar to \cite{linnerAC}:
\begin{align*}
\mathbf{A}_{\boldsymbol{\theta}_{t}}\triangleq & \mathbb{E}_{p_{t}}\Bigl[\nabla_{\boldsymbol{\omega}}f\bigl(\boldsymbol{\omega}_{0};\boldsymbol{s}_{t},\boldsymbol{a}_{t}\bigr)\Bigl(\nabla_{\boldsymbol{\omega}}f\bigl(\boldsymbol{\omega}_{0};\boldsymbol{s}_{t},\boldsymbol{a}_{t}\bigr)\\
 & -\nabla_{\boldsymbol{\omega}}f\bigl(\boldsymbol{\omega}_{0};\boldsymbol{s}_{t+1},\boldsymbol{a}_{t+1}'\bigr)\Bigr)^{\intercal}\Bigr],
\end{align*}
\end{definition}$\mathcal{\hat{F}}$ is a sufficiently rich function
class for a large critic network width $m_{Q}$ and radius $R_{\boldsymbol{\omega}}$.
It's worth noting that function $\hat{f}\in\mathcal{\hat{F}}$ can
be seen as a local linearization of the critic DNN function $f$ and
satisfies a nice property 
\begin{equation}
\hat{f}\bigl(\boldsymbol{\omega}_{a}\bigr)-\hat{f}\bigl(\boldsymbol{\omega}_{b}\bigr)=\bigl\langle\nabla_{\boldsymbol{\omega}}f\bigl(\boldsymbol{\omega}_{0}\bigr),\boldsymbol{\omega}_{a}-\boldsymbol{\omega}_{b}\bigr\rangle.\label{eq:linear property}
\end{equation}
By introducing auxiliary function $\hat{f}$, the error $\epsilon_{\mathrm{cri}}\bigl(t\bigr)$
is decomposed into two parts, i.e., the local linearization error
between $f$ and $\hat{f}$ in (\ref{eq:bias1}) and the bias $\bigl|\mathbb{E}\bigl[\hat{f}\bigl(\bar{\boldsymbol{\omega}}_{t}^{i}\bigr)\bigr]-\hat{Q}_{i}^{\pi_{\boldsymbol{\theta}_{t}}}\bigr|,\forall i$.
Specifically, we give the linearization error in Appendix A and focus
on deriving $\bigl|\mathbb{E}\bigl[\hat{f}\bigl(\bar{\boldsymbol{\omega}}_{t}^{i}\bigr)\bigr]-\hat{Q}_{i}^{\pi_{\boldsymbol{\theta}_{t}}}\bigr|,\forall i$,
in the following.

\subsubsection{Assumptions on the Targets}

For the surrogate Q functions $\bigl\{\hat{Q}_{i}^{\pi_{\boldsymbol{\theta}_{t}}},\forall i\bigr\}$,
we have the following standard assumptions:

\begin{assumption}\rm \textit{(Assumptions on $\bigl\{\hat{Q}_{i}^{\pi_{\boldsymbol{\theta}_{t}}}\bigr\}_{i=0,,\ldots I}$:)}\\
1) The inequality $\boldsymbol{\omega}^{\intercal}\mathbf{A}_{\boldsymbol{\theta}_{t}}\boldsymbol{\omega}>0,\forall t$
holds for any $\boldsymbol{\omega}$, which further implies that there
is a lower bound $\varsigma>0$, such that
\[
\lambda_{\mathrm{min}}\left(\mathbf{A}_{\boldsymbol{\theta}_{t}}+\mathbf{A}_{\boldsymbol{\theta}_{t}}^{\intercal}\right)\geq\varsigma
\]
holds uniformly for all $\boldsymbol{\theta}_{t}\in\Theta$, where
$\lambda_{\mathrm{min}}\bigl(\cdot\bigr)$ represents the smallest
eigenvalue. \linebreak{}
2) $\mathcal{\hat{F}}$ is closed under the Bellman operator, and
there is a point $\dot{\boldsymbol{\omega}}_{t}^{i}$ in the constraint
set $\mathbf{\Omega}_{i}=\mathbb{B}\left(\boldsymbol{\omega}_{0}^{i},R_{w}\right)$
such that $\hat{f}\left(\dot{\boldsymbol{\omega}}_{t}^{i}\right)=\mathcal{T}_{\pi_{\boldsymbol{\theta}_{t}}}\hat{f}\left(\dot{\boldsymbol{\omega}}_{t}^{i}\right)=\hat{Q}_{i}^{\pi_{\boldsymbol{\theta}_{t}}},\forall i$
for any $\pi_{\boldsymbol{\theta}_{t}}$.\end{assumption}

Note that Assumption 3.1 is standard as in \cite{linnerAC} to guarantee
the existence and uniqueness of the problem (\ref{eq:MSBE}). Assumption
3.2 is a regularity condition commonly used in \cite{2019b}, \cite{Assumption31}
and \cite{Assumption32}, which states that the representation power
of $\mathcal{\hat{F}}$ is sufficiently rich to represent $\hat{Q}_{i}^{\pi_{\boldsymbol{\theta}_{t}}},\forall i$.
Based Assumption 3.2 and the property (\ref{eq:linear property}),
it is obtained that, $\forall i$,
\begin{align}
 & \bigl|\mathbb{E}\bigl[\hat{f}\bigl(\bar{\boldsymbol{\omega}}_{t}^{i}\bigr)\bigr]-\hat{Q}_{i}^{\pi_{\boldsymbol{\theta}_{t}}}\bigl|\label{eq:error1}\\
\leq & \bigl\Vert\nabla_{\boldsymbol{\omega}}f\bigl(\boldsymbol{\omega}_{0}\bigr)\bigr\Vert_{2}\cdot\bigl\Vert\mathbb{E}\bigl[\bar{\boldsymbol{\omega}}_{t}^{i}\bigr]-\dot{\boldsymbol{\omega}}_{t}^{i}\bigr\Vert_{2},\forall i.\nonumber 
\end{align}

\subsubsection{Auxiliary Parameters}

For (\ref{eq:error1}), $\bigl\Vert\nabla_{\boldsymbol{\omega}}f\bigl(\boldsymbol{\omega}_{0}\bigr)\bigr\Vert$
can be easily bounded according to Appendix A, however, it is still
non-trivial to derive $\bigl\Vert\mathbb{E}\bigl[\bar{\boldsymbol{\omega}}_{t}^{i}\bigr]-\dot{\boldsymbol{\omega}}_{t}^{i}\bigr\Vert$.
In the single-loop framework, the critic module is only allowed to
update DNNs once or for a few finite times in each iteration, to \textquotedbl learn\textquotedbl{}
the constantly changing targets $\bigl\{\hat{Q}_{i}^{\pi_{\boldsymbol{\theta}_{t}}}\bigr\}_{i=0,,\ldots I}$.
Thus, the error between $\mathbb{E}\bigl[\bar{\boldsymbol{\omega}}_{t}^{i}\bigr]$
and $\dot{\boldsymbol{\omega}}_{t}^{i}$ is introduced both by the
finite-time critic update and by the change of policy $\pi_{\boldsymbol{\theta}_{t}}$.

To separate the two coupled errors, we define two auxiliary parameters
$\bigl\{\boldsymbol{m}_{t}^{i},\bar{\boldsymbol{m}}_{t}^{i},\forall i\bigr\}$.
The two auxiliary parameters are updated by the same rules as $\bigl\{\boldsymbol{\omega}^{i},\boldsymbol{\bar{\omega}}^{i},\forall i\bigr\}$
but use the local linearization function $\hat{f}$ and auxiliary
observations $\tilde{\varepsilon}_{t}=\bigl\{\tilde{\boldsymbol{s}}_{t},\tilde{\boldsymbol{a}}_{t},\bigl\{ C_{i}(\tilde{\boldsymbol{s}}_{t},\tilde{\boldsymbol{a}}_{t})\bigr\}_{i=0,\ldots,I},\tilde{\boldsymbol{s}}_{t+1}\bigr\}$,
where $\tilde{\varepsilon}_{t}$ is sampled by $\pi_{\boldsymbol{\theta}_{t}}$
before the $n_{t}$-th iteration and by the fixed policy $\pi_{\boldsymbol{\theta}_{n_{t}+1}}$
after the $(n_{t}+1)$-th iteration, and where we set $n_{t}=t-t^{\kappa_{5}},\kappa_{5}\in(0,1)$.
Please refer to (\ref{eq:m_TD_learning})-(\ref{eq:auxiliary delta})
for details. Based on this, we can decompose the convergence error
into the error caused by the change of policy and the error caused
by finite critic updates. Thus, we have that
\begin{equation}
\bigl\Vert\mathbb{E}\bigl[\bar{\boldsymbol{\omega}}_{t}^{i}\bigr]-\dot{\boldsymbol{\omega}}_{t}^{i}\bigr\Vert\leq\bigl\Vert\mathbb{E}\bigl[\bar{\boldsymbol{\omega}}_{t}^{i}\bigr]-\mathbb{E}\bigl[\bar{\boldsymbol{m}}_{t}^{i}\bigr]\bigr\Vert+\bigl\Vert\mathbb{E}\bigl[\bar{\boldsymbol{m}}_{t}^{i}\bigr]-\dot{\boldsymbol{\omega}}_{t}^{i}\bigr\Vert,
\end{equation}
$,\forall i$, where the first term presents the distance between
the fixed policy update trajectory and the unfixed policy update trajectory,
and the second term characterizes the error induced by finite-time
critic updates. We bounded them in Appendix B, respectively.

\subsubsection{The Convergence Rate of Critic Module}

Finally, we obtain the convergence rate of the critic module:

\noindent \newtheorem{lemma}{Lemma}

\begin{lemma} \textit{(Convergence rate of the Critic Module)} \\
Suppose Assumption 1 and Assumption 3 hold, the width of critic DNNs
is $m_{Q}$, and the radius $R_{\boldsymbol{\omega}}$ of the parameter
constraint set $\mathbf{\Omega}=\mathbb{B}\left(\boldsymbol{\omega}_{0}^{i},R_{\boldsymbol{\omega}}\right)$
is specifically set to $R_{\boldsymbol{\omega}}=a_{0}m_{Q}^{-1/2}L^{-4/9}$.
Then, it holds that
\begin{align*}
\epsilon_{\mathrm{cri}}\bigl(t\bigr)\triangleq & \Bigl|\mathbb{E}\bigl[f\bigl(\bar{\boldsymbol{\omega}}_{t}^{i};\boldsymbol{s},\boldsymbol{a}\bigr)\bigr]-\hat{Q}_{i}^{\pi_{\boldsymbol{\theta}_{t}}}\bigl(\boldsymbol{s},\boldsymbol{a}\bigr)\Bigl|\\
\leq & O\Bigl(\epsilon_{m_{Q}}+\frac{\bigl(1-\gamma_{t}\bigr)^{t^{\kappa_{5}}/2}}{\gamma_{t}^{1/2}}+m_{Q}\gamma_{n_{t}}^{1/2}\eta_{n_{t}}^{1/2}t^{0.215}\\
 & +m_{Q}\eta_{n_{t}}t^{-0.57}+\frac{\gamma_{n_{t}}^{1/2}}{\eta_{n_{t}}^{1/2}}+m_{Q}\eta_{n_{t}}\beta_{n_{t}}t^{0.86}\Bigr),
\end{align*}
with almost probability 1, where $\epsilon_{m_{Q}}\triangleq O\bigl(m_{Q}^{-1/6}\sqrt{\mathrm{log}\,m_{Q}}\bigr)$
and $\underset{m_{Q}\rightarrow\infty}{\mathrm{lim}}\epsilon_{m_{Q}}=0$.

\end{lemma}

Remark that we choose this particular radius $R_{\boldsymbol{\omega}}$
= $a_{0}m_{Q}^{-1/2}L^{-4/9}$ following the same setup as \cite{DQlearning}
just for tractable convergence analysis. Although $R_{\boldsymbol{\omega}}$
in Lemma 1 seems small but is in fact sufficiently large to enable
a powerful representation capability for the critic DNNs to fit the
training data because the weights are randomly initialized (per entry)
around $m_{Q}^{-1/2}$ for $m_{Q}$ being large \cite{Allen2019b}.
In particular, \cite{Allen2019b} states that with the DNN defined
in \ref{subsec:Neural-Network-Parametrization}, the SGD method can
find global minima in the neighborhood $\mathbb{B}\left(\boldsymbol{\omega},R_{\boldsymbol{\omega}}\right)$
on the training objective of overparameterized DNNs, as long as $R_{\boldsymbol{\omega}}\geq O\bigl(m_{Q}^{-1/2}\bigr)$.

According to Assumption 2 and 
\begin{equation}
\sum_{t=0}^{\infty}\alpha_{t}\bigl|f\bigl(\bar{\boldsymbol{\omega}}_{t}^{i};\boldsymbol{s},\boldsymbol{a}\bigr)-\hat{Q}_{i}^{\pi_{\boldsymbol{\theta}_{t}}}\bigl(\boldsymbol{s},\boldsymbol{a}\bigr)\bigl|<\infty,
\end{equation}
which is required in the proof of asymptotic consistency of surrogate
functions below (Please refer to (\ref{eq:stochastic bias}) for details).
More intuitively, if we set the step size $\alpha_{t}=O\bigl(m_{Q}^{-1/2}t^{-\kappa_{1}}\bigr)$,
we can further have
\begin{align}
\epsilon_{\mathrm{cri}}\bigl(t\bigr)\leq & O\bigl(t^{\kappa_{1}-1}\bigr).\label{eq:critic_convergence_rate1}
\end{align}

\subsection{Consistency of Surrogate Functions\label{subsec:Asymptotic-Consistency}}

Although the single-loop design and the observation reuse make $\tilde{J}_{i}^{t}$
and $\tilde{\boldsymbol{g}}_{i}^{t}$ biased estimation, we manage
to prove that $\hat{J}_{i}^{t}$ and $\hat{\boldsymbol{g}}_{i}^{t}$
used to construct the surrogate functions satisfy asymptotic consistency,
which is a key to establish the global convergence:

\begin{lemma} \textit{(Asymptotic consistency of surrogate functions)}\\
Suppose that Assumptions 1-3 are satisfied, and the width of critic
DNNs is $m_{Q}$. Then, for all $i\in\left\{ 0,1,\ldots,I\right)$,
we have
\begin{align}
\underset{t\rightarrow\infty}{\mathrm{lim}} & \bigl|\hat{J}_{i}^{t}-J_{i}\left(\boldsymbol{\theta}_{t}\right)\bigl|=0,\label{eq:JJ}\\
\underset{t\rightarrow\infty}{\mathrm{lim}} & \bigl\Vert\hat{\boldsymbol{g}}_{i}^{t}-\nabla J_{i}\left(\boldsymbol{\theta}_{t}\right)\bigl\Vert_{2}\leq\epsilon_{m_{Q}},\label{eq:gg}
\end{align}
where $\epsilon_{m_{Q}}\triangleq O\bigl(m_{Q}^{-1/6}\sqrt{\mathrm{log}\,m_{Q}}\bigr)$
and $\underset{m_{Q}\rightarrow\infty}{\mathrm{lim}}\epsilon_{m_{Q}}=0$.\end{lemma}
\begin{IEEEproof}
Since we adopt recursive operations to obtain $\hat{J}_{i}^{t}$ and
$\hat{\boldsymbol{g}}_{i}^{t}$ in (\ref{eq:J-average}) and (\ref{eq:g-average}),
the proof can be completed as long as some key conditions are proved
to be satisfied, i.e., $\sum_{t=0}^{\infty}\alpha_{t}\bigl|\mathbb{E}\bigl[\tilde{J}_{i}^{t}\bigr]-J_{i}\bigl(\boldsymbol{\theta}_{t}\bigr)\bigr|\text{<}\infty$
and $\sum_{t=0}^{\infty}\alpha_{t}\bigl\Vert\mathbb{E}\bigl[\tilde{\boldsymbol{g}}_{i}^{t}\bigr]-\nabla J_{i}\left(\boldsymbol{\theta}_{t}\right)\bigr\Vert_{2}\text{<}\infty$
according to Lemma 5. Please refer to Appendix C for details.
\end{IEEEproof}

\subsection{Convergence of the Actor Module\label{subsec:Convergence-of-Actor}}

This subsection proves that the proposed SLDAC algorithm converges
to a KKT point (up to an error of $\epsilon_{m_{Q}}$ with $\underset{m_{Q}\rightarrow\infty}{\mathrm{lim}}\epsilon_{m_{Q}}=0$)
of the original Problem (1). We begin with the definition of KKT solutions
of the original problem.\begin{definition} \textit{(KKT solution
of the original problem) }\\
A solution $\boldsymbol{\theta}^{*}$ is called a KKT soluction of
the original problem, it there exist Lagrange multipliers $\mathbf{\lambda}=\bigl[\lambda_{1},\ldots,\lambda_{I}\bigr]^{\intercal}\succeq\mathbf{0}$,
such that the following conditions are satisfied:
\begin{equation}
\boldsymbol{g}_{0}\left(\boldsymbol{\theta}^{*}\right)+\sum_{i}\lambda_{i}\boldsymbol{g}_{i}\left(\boldsymbol{\theta}^{*}\right)=\mathbf{0}\label{eq:condition1}
\end{equation}
\begin{equation}
J_{i}\left(\boldsymbol{\theta}^{*}\right)\leq0,i=1,\ldots,I,\label{eq:condition2}
\end{equation}
\begin{equation}
\lambda_{i}J_{i}\left(\boldsymbol{\theta}^{*}\right)=0,i=1,\ldots,I.\label{eq:condition3}
\end{equation}
\end{definition}

Moreover, considering a subsequence $\bigl\{\boldsymbol{\theta}_{t_{j}}\bigr\}_{j=1}^{\infty}$converging
to a limiting point $\boldsymbol{\theta}^{*}$, there exist converged
surrogate functions $\hat{J}_{i}\left(\boldsymbol{\theta}\right),\forall i$
such that
\begin{equation}
\underset{j\rightarrow\infty}{\mathrm{lim}}\bar{J}_{i}^{t_{j}}\left(\boldsymbol{\theta}\right)=\hat{J}_{i}\left(\boldsymbol{\theta}\right),\forall\boldsymbol{\theta}\in\mathbf{\Theta},\label{eq:converged surrogate function}
\end{equation}
where
\begin{equation}
\bigl|\hat{J}_{i}\left(\boldsymbol{\theta}^{*}\right)-J_{i}\left(\boldsymbol{\theta}^{*}\right)\bigl|=0,
\end{equation}
\begin{equation}
\bigl\Vert\nabla\hat{J}_{i}\left(\boldsymbol{\theta}^{*}\right)-\nabla J_{i}\left(\boldsymbol{\theta}^{*}\right)\bigl\Vert_{2}=0.
\end{equation}
Then, with Lemma 2 and Assumptions 1 - 3, we are ready to prove the
main convergence theorem:

\noindent \newtheorem{theorem}{Theorem}

\begin{theorem} \textit{(Global Convergence of Algorithm 1:) }\\
Suppose Assumptions 1 - 3 are satisfied, the initial point $\boldsymbol{\theta}_{0}$
is feasible,  i.e.,$\textrm{max}_{i\in\left\{ 1,\ldots,I\right\} }J_{i}\left(\boldsymbol{\theta}_{0}\right)\leq0$,
and the number of data samples is set to $T_{t}=O\left(\textrm{log}t\right)$.
Denote $\left\{ \boldsymbol{\theta}_{t}\right\} _{t=1}^{\infty}$
as the iterates generated by Algorithm 1 with a sufficiently small
initial step size $\beta_{0}$. Then, every limiting point $\boldsymbol{\theta}^{*}$
of $\left\{ \boldsymbol{\theta}_{t}\right\} _{t=1}^{\infty}$ satisfying
the Slater condition satisfies the KKT conditions in (\ref{eq:condition1}),
(\ref{eq:condition2}), and (\ref{eq:condition3}) up to an error
of $\varepsilon_{m_{Q}}\triangleq O\bigl(m_{Q}^{-1/6}\sqrt{\mathrm{log}\,m_{Q}}\bigr)$,
i.e.,
\begin{equation}
\bigl\Vert g_{0}\left(\boldsymbol{\theta}^{*}\right)+\sum_{i}\lambda_{i}g_{i}\left(\boldsymbol{\theta}^{*}\right)\bigl\Vert_{2}\leq\epsilon_{m_{Q}}\label{eq:condition1-1}
\end{equation}
\begin{equation}
J_{i}\left(\boldsymbol{\theta}^{*}\right)\leq\epsilon_{m_{Q}},i=1,\ldots,I,\label{eq:condition2-1}
\end{equation}
\begin{equation}
\lambda_{i}J_{i}\left(\boldsymbol{\theta}^{*}\right)\leq\epsilon_{m_{Q}},i=1,\ldots,I.\label{eq:condition3-1}
\end{equation}
where $\underset{m_{Q}\rightarrow\infty}{\mathrm{lim}}\epsilon_{m_{Q}}=0$.\end{theorem}
According to this theorem, the reader can appropriately take $m_{Q}$
that satisfies the performance requirements. The key challenges in
the convergence analysis lie in the proof of Lemma 1-3. Once Lemma
1-3 are proved in this paper, Theorem 1 follows from the similar analyses
in our previous work \cite{Yangrui} and \cite{SCAOPO}, and we omit
it due to the space limit. 

Finally, we discuss the convergence behavior of Algorithm 1 with an
infeasible initial point. In this case, it follows from the same analysis
in Appendix B of our related work \cite{SCAOPO} that Algorithm 1
either converges to a KKT point of Problem (\ref{eq:problem 1}),
or converges to the following \textit{undesired set}:
\begin{equation}
\overline{\mathbf{\Theta}}_{C}^{*}=\left\{ \boldsymbol{\theta}:J\left(\boldsymbol{\theta}\right)>0,\boldsymbol{\theta}\in\mathbf{\Theta}_{C}^{*}\right\} ,
\end{equation}
where $\mathbf{\Theta}_{C}^{*}$ is the set of stationary points of
the following constraint minimization problem:
\begin{equation}
\underset{\boldsymbol{\theta}\in\mathbf{\Theta}}{\textrm{min}}J\left(\boldsymbol{\theta}\right)\stackrel{\Delta}{=}\textrm{max}_{i\in\left\{ 1,\ldots,I\right\} }J_{i}\left(\boldsymbol{\theta}\right)
\end{equation}
Thanks to the proposed feasible update (\ref{eq:feasible update}),
Algorithm 1 may still converge to a KKT point (up to an error of $\epsilon_{m_{Q}}$)
of Problem (\ref{eq:problem 1}) even with an infeasible initial point,
as long as the initial point is not close to an undesired point $\boldsymbol{\theta}_{C}^{*}\in\overline{\mathbf{\Theta}}_{C}^{*}$
such that the algorithm gets stuck in this undesired point. In practice,
if we generate multiple random initial points, it is likely that one
of the iterates starting from these random initializers will not get
stuck in undesired points, and the algorithm will converge to a KKT
point (up to an error of $\epsilon_{m_{Q}}$) of Problem (\ref{eq:problem 1}).

\section{Simulation Results}

In this section, we apply the proposed SLDAC to solve the three typical
CRL application problems in Section \ref{subsec:Problem-Formulation}.
We maintain the lengths of observation storage $T=1000\text{,30000,500}$
and interact with the environment to obtain $B=100,10000,100$ new
observations at each iteration in Example 1, Example 2, and Example
3, respectively. The policy mean network and the standard deviation
network, as well as all critic networks of all algorithms are parameterized
by the fully-connected DNN defined in Section \ref{subsec:Neural-Network-Parametrization},
where the number of hidden layers is $2$ and the width is $128$.
In addition, we updated the critic parameters $\boldsymbol{\omega}_{i},\forall i$
using a fixed step size $\eta_{t}=0.001$, and then fine tune the
other step sizes $\bigl\{\alpha_{t},\beta_{t},\gamma_{t}\bigr\}$
to achieve a good convergence speed. Please note that although we
appropriately relax some conditions assumed for rigorous convergence
proof, i.e., $T_{t}=O\left(\textrm{log}t\right)$, $m_{Q}$ is sufficiently
large and the requirements for step sizes given in Assumption 2, the
algorithm can still have a good convergence behavior.

We adopt the advanced Actor-only algorithm SCAOPO \cite{SCAOPO} and
the classical two-loop DAC algorithms, PPO-Lag \cite{PPOLagTRPOLag}
and CPO \cite{CPO}, as baselines to evaluate the performance of dual
critic DNNs and the CSSCA-based policy optimization of the proposed
SLDAC, respectively. To demonstrate the effect of the single-loop
design, we simulate the SLDAC with $q=1,5$, and 10 critic updates
at each iteration, respectively. Since the two-loop DAC requires the
critic module to constantly update until the Q-values are accurately
estimated, which is impractical, so we did not directly show it in
the simulation results. However, the proposed algorithm with a larger
$q$, i.e., $q=10$ in Example 1 and 3, and $q=50$ in Example 2,
can be viewed as a practical implementation of the two-loop DAC. We
also simulate the SLDAC without storing and reusing previous data
samples to demonstrate the benefit of reusing old experiences. Moreover,
the results of each algorithm are averaged across 10 random seeds.
\begin{figure*}
\begin{centering}
\subfloat[\textcolor{blue}{\label{fig:MIMO}}Delay-constrained power control
for downlink MU-MIMO.]{\centering{}\includegraphics[width=6cm,height=9cm]{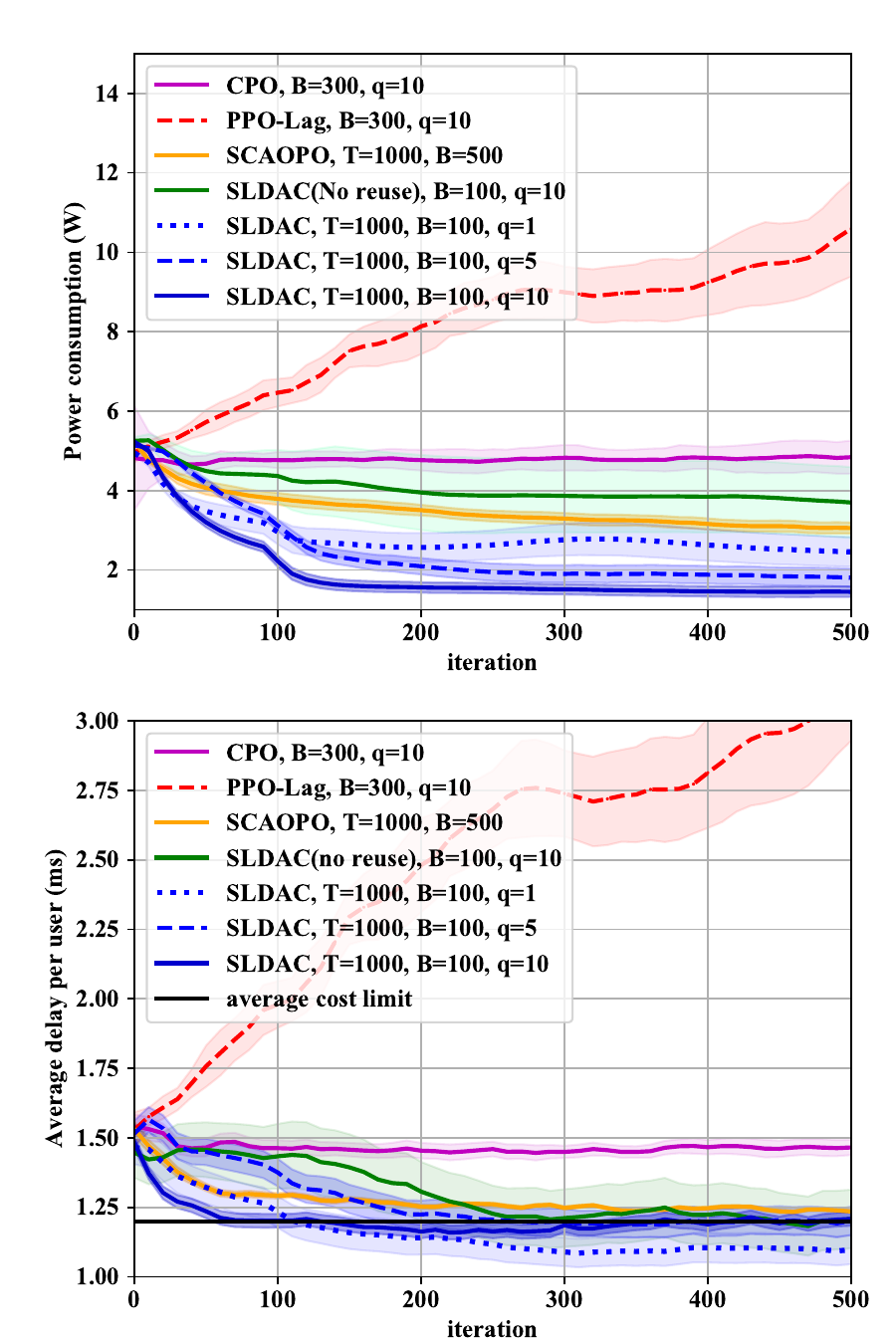}}\subfloat[\textcolor{blue}{\label{fig:Safety Gym}}Autonomous vehicle transport
with safety assurance.]{\begin{centering}
\includegraphics[width=6cm,height=9cm]{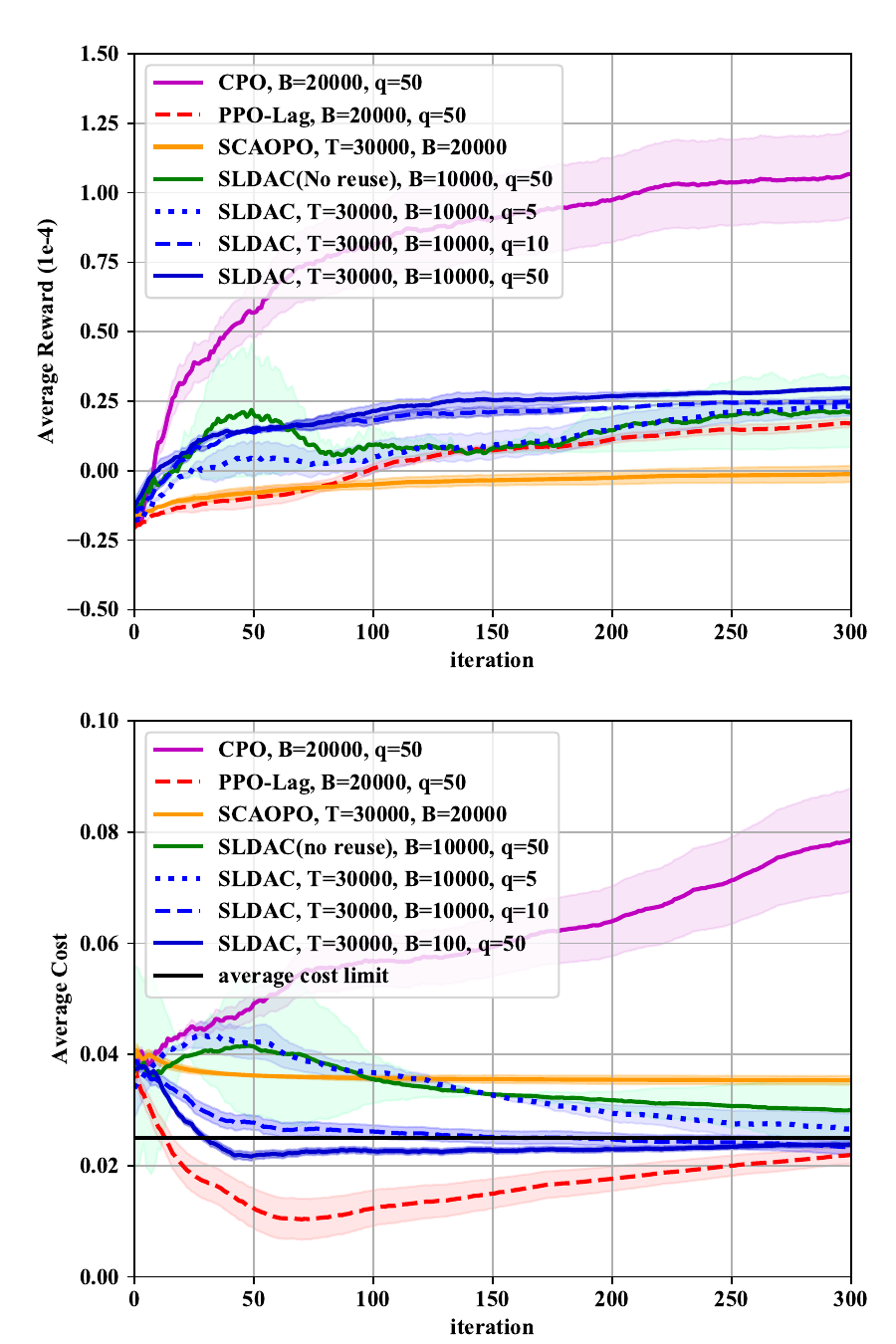}
\par\end{centering}
}\subfloat[\textcolor{blue}{\label{fig:CLQR}}Constrained linear-quadratic regulator.]{\begin{centering}
\includegraphics[width=6cm,height=9cm]{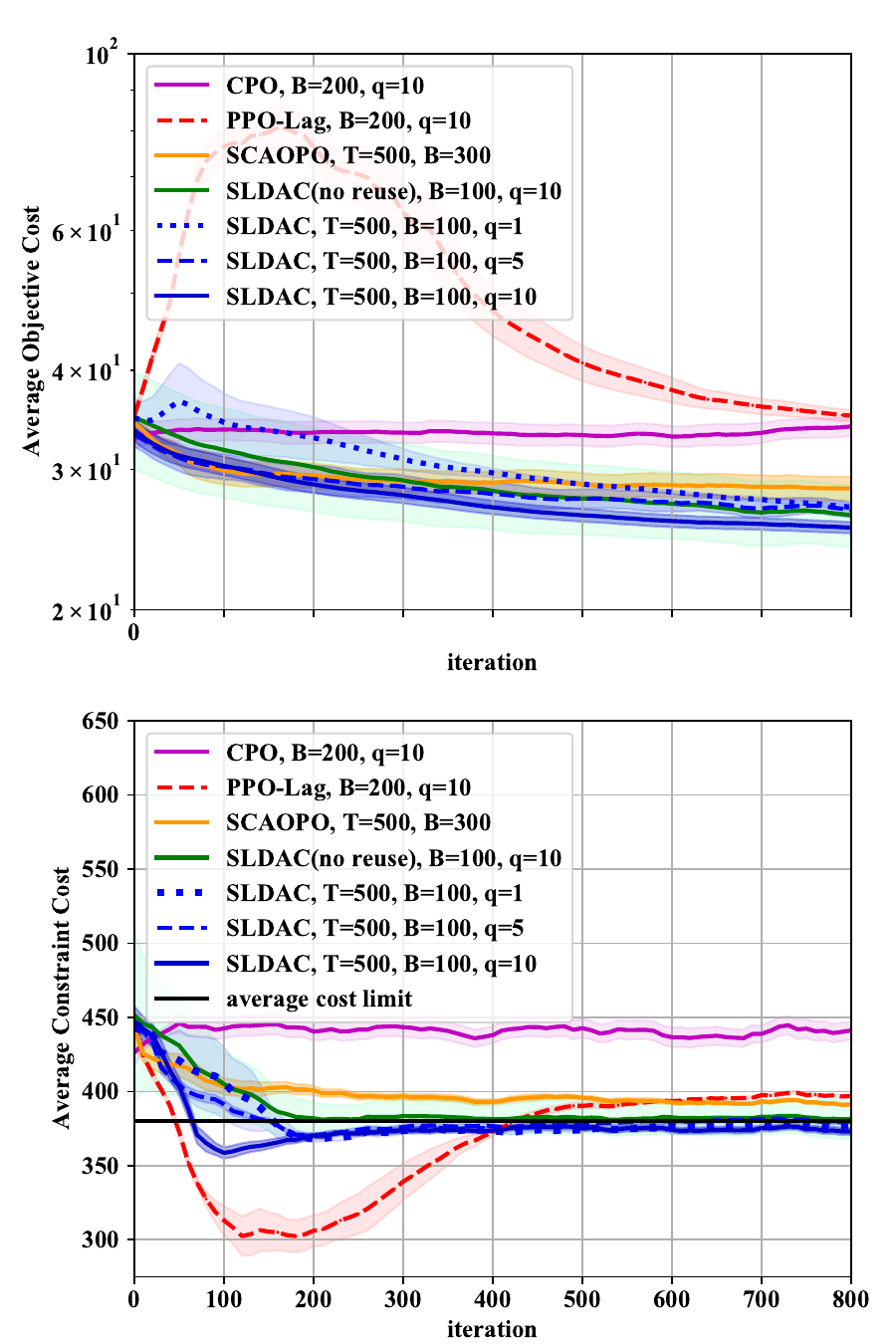}
\par\end{centering}
}
\par\end{centering}
\caption{\label{fig:example1}The first and second rows of images respectively
show the learning curves of average rewards and costs in three typical
scenarios, respectively, where the lines show the average performance
of the algorithms, and the shade regions indicate half the standard
deviations.}
\vspace{-0.5cm}
\end{figure*}

\subsection{Delay-Constrained Power Control for Downlink MU-MIMO\label{subsec:case1}}

Following the same standard setup implemented as \cite{SCAOPO}, we
adopt a geometry-based channel model $\boldsymbol{h}_{k}=\sum_{i=1}^{N_{p}}\bar{\alpha}_{k,i}\mathbf{a}\bigl(\psi_{k,i}\bigr)$
for simulation, where $\mathbf{a}\bigl(\psi_{k,i}\bigr)=\bigl[1,e^{j\pi sin\left(\psi\right)},\ldots,e^{j\bigl(N_{t}-1\bigr)\pi sin\left(\psi\right)}\bigr]^{\intercal}$
is the half-wavelength spaced uniform linear array (ULA) response
vector, $N_{p}$ denotes the number of scattering path, and $\psi_{k,i}$denotes
the $i$-th angle of departure (AoD). We assume that $\varphi_{k,i}$'s
with an angular spread $\sigma_{AS}=5$ and $\bar{\alpha}_{k,i}\sim\mathcal{CN}\bigl(0,\bar{\sigma}_{k,i}^{2}\bigr)$
are Laplacian distributed, $\bar{\sigma}_{k,i}^{2}$'s follow an exponential
distribution and are normalized such that $\sum_{i=1}^{N_{p}}\bar{\sigma}_{k,i}^{2}=g_{k}$,
where $g_{k}$ represents the path gain of the $k$-th user. Specially,
we uniformly generate the path gains $g_{k}$'s from -10 dB to 10
dB and set $N_{p}=4$ for each user. In addition, the bandwidth $\bar{B}=10$
MHz, the duration of one time slot $\bar{\tau}=1$ ms, the noise power
density $-100$ dBm/Hz, and the arrival data rate $A_{k},\forall k$
are uniformly distributed over $\bigl[0,20\bigr]$ Mbit/s. The constants
in the surrogate problems are $\zeta_{k}=1,\forall k$. In this case,
we choose the step sizes as $\alpha_{t}=\frac{1}{t^{0.6}}$, $\beta_{t}=\frac{1}{t^{0.7}}$,
and $\gamma_{t}=\frac{1}{t^{0.3}}$.\lyxdeleted{86195}{Sun May 26 00:58:17 2024}{ }

In Fig. \ref{fig:example1}, we plot the average power consumption
and the average delay per user when the number of transmitting antennas
$N_{t}=8$ and the number of receiving antennas for users $K=4$.
Compared with the classical DAC algorithms PPO-Lag and CPO, the proposed
SLDAC can significantly reduce power consumption while meeting the
delay constraint regardless of whether data reuse is adopted, which
reveals the benefit of the guarantee of convergence to a KKT point.
In terms of SCAOPO, its convergence rate depends heavily on the number
of observations, because its policy gradient is calculated using MC
methods. Even though we feed the SCAOPO more newly added observations
at each iteration, $i.e.,B=500$, the proposed SLDAC still demonstrates
comparable or superior performance to the SCAOPO. For the proposed
SLDAC, it can be seen that the performance at $q=1$ is already good,
the performance at $q=5$ is almost the same as the performance at
$q=10$, and obviously, there is no need to continue increasing $q$.
By comparing the simulation results for $q=5$ and $q=10$, it is
clear that the proposed algorithm with a relatively small $q$ requires
much fewer interactions with the environment to achieve the same convergence
performance, which implies that the single-loop framework can indeed
reduce the interaction cost compared to the two-loop framework. In
addition, the SLDAC with observation reuse can attain better performance
than that without observation because the old observations generated
from the old policy also contain information that can be exploited. 

\subsection{Robot Navigation with Safety Assurance}

By choosing the step sizes $\alpha_{t}=\frac{1}{t^{0.55}}$, $\beta_{t}=\frac{1}{t^{0.75}}$,
and $\gamma_{t}=\frac{1}{t^{0.4}}$ and setting the average cost limit
$c_{1}=0.025$ and the surrogate problems are $\zeta_{i}=1,\forall i$,
we obtain the simulation results shown in Fig. \ref{fig:Safety Gym}.
It can be seen that CPO failed due to significant approximation errors
in this complex scenario. Although PPO-Lag was able to meet the constraints,
the average reward obtained is relatively low due to its simple stochastic
gradient descent-based policy update method. In terms of SCAOPO, it
can only receive a negative average reward because the MC method has
a large error in estimating Q in this scenario. In contrast, the proposed
SLDAC can receive significantly higher average reward while meeting
the average cost limit regardless of whether data reuse is adopted,
which reveals the benefit of the guarantee of convergence to a KKT
point. Moreover, due to the special single-loop framework design,
the SLDAC algorithm has a good performance even with a relatively
small $q$ than that in baseline algorithms.

\subsection{Constrained Linear-Quadratic Regulator}

Similar to \cite{CLQset} and \cite{SCAOPO}, we set the state dimension
$n_{\boldsymbol{s}}=15$, action dimension $n_{\boldsymbol{a}}=4$,
the average constraint cost limit $c_{1}=380$, and the constants
in the surrogate problems $\zeta_{i}=10,\forall i$. Then, by choosing
the step sizes $\alpha_{t}=\frac{1}{t^{0.6}}$, $\beta_{t}=\frac{1}{t^{0.8}}$,
and $\gamma_{t}=\frac{1}{t^{0.27}}$, we obtain the simulation results
shown in Fig. \ref{fig:CLQR}. As it can be seen from the figures,
we obtain a similar behavior as in the other two scenario that the
proposed SLDAC with appropriately chosen $q$ provides the best performance.

\section{Conclusion}

In this work, we proposed a novel SLDAC algorithm, which is the first
single-loop DAC variant to solve general CMDPs with guaranteed convergence.
Compared with existing algorithms, the proposed SLDAC considers more
requirements in many realistic domains. Specifically, the CSSCA-based
policy optimization method is used to better handle the non-convex
stochastic objective and constraints, and the single-loop framework
and observation reuse are adopted to reduce the computational complexity
and interaction cost. Under some technical conditions, we provide
the finite-time convergence rate of critic DNNs and the asymptotic
consistency of estimated surrogate functions. Finally, we prove that
the SLDAC converges to a KKT point of the original problem almost
surely. Simulation results show that the proposed SLDAC can achieve
a better overall performance compared to baselines with much lower
interaction cost.

\begin{appendices}

\section{Technical Bounds about DNNs}

In this section, we show some bounds for the critic DNNs and the error
introduced by local linearization. Recalling that $m_{Q}$ represents
the width of critic DNNs, $L$ stands for the depth of the critic
DNNs, and $R_{\boldsymbol{\omega}}$ denotes the radius of the constraint
set $\mathbf{\Omega}=\mathbb{B}\left(\boldsymbol{\omega}_{0}^{i},R_{\boldsymbol{\omega}}\right)$
for the parameter $\boldsymbol{\omega}$, we restate a useful lemma
from recent studies of overparameterized DNNs \cite[Lemma 6.4]{DQlearning},
\cite[Lemma B.3]{Cao2019b}, and \cite[Theorem 5]{Allen2019b}.\begin{lemma}
\textit{(}\textsl{Technical Bounds about DNN}:\textit{)}\\
Let $\sigma\in\bigl(0,1\bigr)$, $d$ and $\left\{ a_{i}\right\} _{i=0,1,\ldots}$
denote some universal constants that are independent of problem parameters
throughout this paper. Then, for any $\sigma>0$, if $a_{1}d^{3/2}L^{-1}m_{Q}^{-3/4}\leq R_{\boldsymbol{\omega}}\leq a_{2}L^{-6}\bigl(\mathrm{log}\,m_{Q}\bigr)^{-3}$
and $m_{Q}\geq a_{3}\mathrm{max}\bigl\{ dL^{2}\mathrm{log}\left(m_{Q}/\sigma\right),R_{\boldsymbol{\omega}}^{-4/3}L^{-8/3}\mathrm{log}\left(m_{Q}/\left(R_{\boldsymbol{\omega}}\sigma\right)\right)\bigr\}$
are satisfied, it holds that the difference between the $f\left(\boldsymbol{\omega}\right)$
and its local linearization $\hat{f}\left(\boldsymbol{\omega}\right)$
\[
\bigl|f\left(\boldsymbol{\omega};\boldsymbol{s},\boldsymbol{a}\right)-\hat{f}\left(\boldsymbol{\omega};\boldsymbol{s},\boldsymbol{a}\right)\bigr|\leq a_{4}R_{\boldsymbol{\omega}}^{4/3}L^{4}\sqrt{m_{Q}\mathrm{log}\,m_{Q}}+a_{5}R_{\boldsymbol{\omega}}^{2}L^{5},
\]
$\forall\boldsymbol{\phi}\bigl(\boldsymbol{s},\boldsymbol{a}\bigr)\in\mathbb{R}^{d}$,
with probability at least $1-\sigma-\mathrm{exp}\bigl\{-a_{6}m_{Q}R_{\boldsymbol{\omega}}^{2/3}L\bigr\}$.
Moreover, the gradient of the DNN is also bounded as $\bigl\Vert\nabla_{\boldsymbol{\omega}}f\left(\boldsymbol{\omega};\boldsymbol{s},\boldsymbol{a}\right)\bigr\Vert_{2}\leq a_{7}m_{Q}^{1/2}$
with probability at least $1-L^{2}\mathrm{exp}\bigl\{-a_{8}m_{Q}R_{\boldsymbol{\omega}}^{2/3}L\bigr\}$.\end{lemma} 

Specially, we set all critic parameters to have the same initial values,
i.e., $\boldsymbol{\omega}_{0}^{i}=\boldsymbol{\bar{\omega}}_{0}^{i}=\boldsymbol{m}_{0}^{i}=\bar{\boldsymbol{m}}_{0}^{i}=\boldsymbol{\omega}_{0},\forall i$,
and generate the initial parameter $\boldsymbol{\omega}_{0}$ from
$\mathcal{N}\left(0,1/m_{Q}^{2}\right)$. Then, it holds that $\bigl\Vert\boldsymbol{\omega}\bigr\Vert_{2}$$\leq a_{9}m_{Q}^{1/2}$.

\section{Proof of Lemma 1}

For simplicity, we omit the superscript $i$ when no confusion arises
in this section. By introducing the auxiliary function class $\mathcal{\hat{F}}$,
the estimated error in the critic module can be divided as follows{\small{}
\begin{align}
 & \bigl|\mathbb{E}\bigl[f\bigl(\bar{\boldsymbol{\omega}}_{t}\bigr)\bigr]-\hat{Q}^{\pi_{\boldsymbol{\theta}_{t}}}\bigr|\label{eq:divide-1}\\
\leq & \underset{\mathrm{bias}_{1}}{\underbrace{\bigl|\mathbb{E}\bigl[f\bigl(\bar{\boldsymbol{\omega}}_{t}\bigr)-\mathbb{E}\bigl[\hat{f}\bigl(\bar{\boldsymbol{\omega}}_{t}\bigr)\bigr]\bigr|}}+\bigl|\mathbb{E}\bigl[\hat{f}\bigl(\bar{\boldsymbol{\omega}}_{t}\bigr)\bigr]-\hat{f}\bigl(\dot{\boldsymbol{\omega}}_{t}\bigr)\bigr|,\nonumber 
\end{align}
}The first term in (\ref{eq:divide-1}), bias 1, characterizes how
far $f$ deviates from its local linearization $\hat{f}$. According
to Appendix A, if we specially set the radius $R_{\boldsymbol{\omega}}=a_{0}m_{Q}^{-1/2}$,
and assume that $m_{Q}$ is sufficiently large, then it holds that
\begin{equation}
\mathrm{bia}_{1}\triangleq\epsilon_{m_{Q}}\leq O\bigl(m_{Q}^{-1/6}\sqrt{\mathrm{log}\,m_{Q}}\bigr).\label{eq:bias1}
\end{equation}
For the second term (\ref{eq:divide-1}), since $\hat{f}$ is a linear
function, we have{\small{}
\begin{align}
\bigl|\mathbb{E}\bigl[\hat{f}\bigl(\bar{\boldsymbol{\omega}}_{t}\bigr)\bigr]-\hat{Q}^{\pi_{\boldsymbol{\theta}_{t}}}\bigl| & \leq\bigl\Vert\nabla_{\boldsymbol{\omega}}f\bigl(\boldsymbol{\omega}_{0}\bigr)\bigr\Vert_{2}\cdot\bigl\Vert\mathbb{E}\bigl[\bar{\boldsymbol{\omega}}_{t}\bigr]-\dot{\boldsymbol{\omega}}_{t}\bigr\Vert_{2},\label{eq:error1-1}
\end{align}
}where recall that $\bigl\Vert\nabla_{\boldsymbol{\omega}}f\left(\boldsymbol{\omega};\boldsymbol{s},\boldsymbol{a}\right)\bigr\Vert_{2}\leq a_{7}m_{Q}^{1/2}$
is given in Appendix A. Then, we analysis the error $\bigl\Vert\mathbb{E}\bigl[\bar{\boldsymbol{\omega}}_{t}\bigr]-\dot{\boldsymbol{\omega}}_{t}\bigr\Vert_{2}$.

We further introduce two auxiliary parameters $\bigl\{\boldsymbol{m},\bar{\boldsymbol{m}},\forall i\bigr\}$,
whose the update process can be formulated as
\begin{equation}
\boldsymbol{m}_{t}=\Pi_{\Omega_{i}}\bigl(\boldsymbol{m}_{t-1}-\eta_{t}\boldsymbol{M}^{\boldsymbol{m}_{t}}\bigr),\forall i,\label{eq:m_TD_learning}
\end{equation}
\begin{equation}
\bar{\boldsymbol{m}}_{t}=\bigl(1-\gamma_{t}\bigr)\bar{\boldsymbol{m}}_{t-1}+\gamma_{t}\boldsymbol{m}_{t}.\label{eq:m^_}
\end{equation}
where recalling $\boldsymbol{\Delta}^{\boldsymbol{\omega}_{t-1}}$
defined in (\ref{eq:gradient term}), the stochastic gradient term
$\boldsymbol{M}^{\boldsymbol{m}_{t-1}}$ is defined as
\begin{align}
\boldsymbol{M}^{\boldsymbol{m}_{t-1}}= & \Bigl(\hat{f}\bigl(\boldsymbol{m}_{t-1};\tilde{\boldsymbol{s}}_{t},\tilde{\boldsymbol{a}}_{t}\bigr)-\bigl(C_{i}^{\text{'}}(\tilde{\boldsymbol{s}}_{t},\tilde{\boldsymbol{a}}_{t})-\hat{J}_{i}^{t-1}\label{eq:auxiliary delta}\\
 & +\hat{f}\bigl(\boldsymbol{m}_{t-1};\tilde{\boldsymbol{s}}_{t+1},\tilde{\boldsymbol{a}}_{t+1}'\bigr)\bigr)\Bigr)\nabla_{\boldsymbol{\omega}}f\left(\boldsymbol{m}_{0};\tilde{\boldsymbol{s}}_{t},\tilde{\boldsymbol{a}}_{t}\right),\nonumber 
\end{align}
where $\tilde{\boldsymbol{a}}_{t+1}'$ is the action chosen by current
policy at the next state $\tilde{\boldsymbol{s}}_{t+1}$. Based on
the auxiliary parameters, we divided $\bigl\Vert\mathbb{E}\bigl[\bar{\boldsymbol{\omega}}_{t}\bigr]-\dot{\boldsymbol{\omega}}_{t}\bigr\Vert_{2}$
into two parts{\small{}
\begin{align}
\bigl\Vert\mathbb{E}\bigl[\bar{\boldsymbol{\omega}}_{t}\bigr]-\dot{\boldsymbol{\omega}}_{t}\bigr\Vert_{2}\leq & \underset{\mathrm{bias}_{2}}{\underbrace{\bigl\Vert\mathbb{E}\bigl[\bar{\boldsymbol{m}}_{t}\bigr]-\dot{\boldsymbol{\omega}}_{t}\bigr\Vert_{2}}}+\underset{\mathrm{bias}_{3}}{\underbrace{\bigl\Vert\mathbb{E}\bigl[\bar{\boldsymbol{\omega}}_{t}\bigr]-\mathbb{E}\bigl[\bar{\boldsymbol{m}}_{t}\bigr]\bigr\Vert_{2}}},\label{eq:divide-2}
\end{align}
}where bias 2 characterizes the error induced by finite-time critic
updates and bias 3 presents the distance between the fixed policy
update trajectory and the unfixed policy update trajectory. We derive
the two in the following, respectively.

\paragraph{bias 2}

According to (\ref{eq:m^_}), we have{\small{}
\begin{equation}
\bar{\boldsymbol{m}}_{t}=\stackrel[t'=0]{t}{\sum}\stackrel[j=t'+1]{t}{\prod}\bigl(1-\gamma_{j}\bigr)\gamma_{t'}\boldsymbol{m}_{t'},\forall i,\label{eq:bias1-1}
\end{equation}
}Noting that $\stackrel[t'=0]{t}{\sum}\stackrel[j=t'+1]{t}{\prod}\left(1-\gamma_{j}\right)\gamma_{t'}=1$,
we can obtain the following derivation according to the Jensen\textquoteright s
inequality:{\small{}
\begin{align}
\bigl\Vert\mathbb{E}\bigl[\bar{\boldsymbol{m}}_{t}\bigr]-\dot{\boldsymbol{\omega}}_{t}\bigl\Vert_{2}^{2}\leq & \stackrel[t'=0]{t}{\sum}\stackrel[j=t'+1]{t}{\prod}\bigl(1-\gamma_{j}\bigr)\gamma_{t'}\bigl\Vert\mathbb{E}\bigl[\boldsymbol{m}_{t'}\bigr]-\dot{\boldsymbol{\omega}}_{t}\bigl\Vert_{2}^{2}\nonumber \\
\leq & \stackrel[t'=0]{t}{\sum}\bigl(1-\gamma_{t}\bigr)^{t-t'}\gamma_{t'}\bigl\Vert\mathbb{E}\bigl[\boldsymbol{m}_{t'}\bigr]-\dot{\boldsymbol{\omega}}_{t}\bigl\Vert_{2}^{2}\nonumber \\
\overset{a}{\leq} & \stackrel[t'=0]{n_{t}}{\sum}\bigl(1-\gamma_{t}\bigr)^{t-t'}\bigl\Vert\mathbb{E}\bigl[\boldsymbol{m}_{t'}\bigr]-\dot{\boldsymbol{\omega}}_{t}\bigl\Vert_{2}^{2}\nonumber \\
 & +\frac{\gamma_{n_{t}}}{\eta_{n_{t}}}\stackrel[t'=n_{t}+1]{t}{\sum}\eta_{t'}\bigl\Vert\mathbb{E}\bigl[\boldsymbol{m}_{t'}\bigr]-\dot{\boldsymbol{\omega}}_{t}\bigl\Vert_{2}^{2}.\label{eq:bias1-2}
\end{align}
}{\small\par}

For the first term of (\ref{eq:bias1-2})-a, since $\mathrm{max}{}_{t'=0,\ldots,n_{t}}\bigl\Vert\mathbb{E}\bigl[\boldsymbol{m}_{t'}\bigr]-\dot{\boldsymbol{\omega}}_{t}\bigl\Vert_{2}^{2}\leq R_{\boldsymbol{\omega}}^{2}=O\bigl(m_{Q}^{-1}\bigr)$,
we can obtain that {\small{}
\begin{align}
\stackrel[t'=0]{n_{t}}{\sum}\bigl(1-\gamma_{t}\bigr)^{t-t'}\bigl\Vert\mathbb{E}\bigl[\boldsymbol{m}_{t'}\bigr]-\dot{\boldsymbol{\omega}}_{t}\bigl\Vert_{2}^{2}\leq & O\bigl(\frac{\bigl(1-\gamma_{t}\bigr)^{t^{\kappa_{5}}}}{\gamma_{t}m_{Q}}\bigr).\label{eq:bias1-3}
\end{align}
}{\small\par}

Now, we derive the second term of (\ref{eq:bias1-2})-a. Recalling
the update rule of (\ref{eq:m_TD_learning}), we have
\begin{align}
\bigl\Vert\mathbb{E}\bigl[\boldsymbol{m}_{t'}\bigr]-\dot{\boldsymbol{\omega}}_{t}\bigl\Vert_{2}^{2} & \leq\bigl\Vert\mathbb{E}\bigl[\Pi_{\Omega_{i}}\bigl(\boldsymbol{m}_{t'}-\eta_{t'}\boldsymbol{M}^{\boldsymbol{m}_{t'}}\bigr)\bigr]-\dot{\boldsymbol{\omega}}_{t}\bigl\Vert_{2}^{2}\nonumber \\
\leq-2\eta_{t'}\bigl\langle\mathbb{E}_{p_{t'}}\bigl[ & \boldsymbol{M}^{\boldsymbol{m}_{t'}}\bigr]-\mathbb{E}_{p_{t'}}\bigl[\boldsymbol{M}^{\dot{\boldsymbol{\omega}}_{t}}\bigr],\mathbb{E}\bigl[\boldsymbol{m}_{t'}\bigr]-\dot{\boldsymbol{\omega}}_{t}\bigr\rangle\nonumber \\
+\bigl\Vert\mathbb{E}\bigl[\boldsymbol{m}_{t'}\bigr]- & \dot{\boldsymbol{\omega}}_{t}\bigl\Vert_{2}^{2}+\eta_{t'}^{2}\bigl\Vert\mathbb{E}\bigl[\boldsymbol{M}^{\boldsymbol{m}_{t'}}\bigr]\bigl\Vert_{2}^{2}.\label{eq:TD-learning decompose}
\end{align}
We first derives the inner product term of (\ref{eq:TD-learning decompose}):
\begin{align}
 & \Bigl\langle\mathbb{E}_{p_{t'}}\bigl[\boldsymbol{M}_{i}^{\boldsymbol{m}_{t'}}\bigr]-\mathbb{E}_{p_{t'}}\bigl[\boldsymbol{M}_{i}^{\dot{\boldsymbol{\omega}}_{t}}\bigr],\mathbb{E}\bigl[\boldsymbol{m}_{t'}\bigr]-\dot{\boldsymbol{\omega}}_{t}\Bigr\rangle\nonumber \\
\overset{a}{=} & \left(\mathbb{E}\bigl[\boldsymbol{m}_{t'}\bigr]-\dot{\boldsymbol{\omega}}_{t}\right)^{\intercal}\mathbf{A}_{\boldsymbol{\theta}_{t}'}\left(\mathbb{E}\bigl[\boldsymbol{m}_{t'}\bigr]-\dot{\boldsymbol{\omega}}_{t}\right)\nonumber \\
= & \frac{1}{2}\left(\mathbb{E}\bigl[\boldsymbol{m}_{t'}\bigr]-\dot{\boldsymbol{\omega}}_{t}\right)^{\intercal}\Bigl(\mathbf{A}_{\boldsymbol{\theta}_{n_{t'}+1}}+\mathbf{A}_{\boldsymbol{\theta}_{n_{t'}+1}}^{\intercal}\Bigr)\left(\mathbb{E}\bigl[\boldsymbol{m}_{t'}\bigr]-\dot{\boldsymbol{\omega}}_{t}\right)\nonumber \\
\geq & \frac{1}{2}\lambda_{\mathrm{min}}\Bigl(\mathbf{A}_{\boldsymbol{\theta}_{n_{t'}+1}}+\mathbf{A}_{\boldsymbol{\theta}_{n_{t'}+1}}^{\intercal}\Bigr)\left\Vert \mathbb{E}\bigl[\boldsymbol{m}_{t'}\bigr]-\dot{\boldsymbol{\omega}}_{t}\right\Vert _{2}^{2}\nonumber \\
\overset{b}{\geq} & \frac{\varsigma}{2}\left\Vert \mathbb{E}\bigl[\boldsymbol{m}_{t'}\bigr]-\dot{\boldsymbol{\omega}}_{t}\right\Vert _{2}^{2},\label{eq:inner product term}
\end{align}
where (\ref{eq:inner product term})-a is according to the linear
property (\ref{eq:linear property}) and (\ref{eq:inner product term})-b
follows Assumption 3-2). Then, by Appendix A and the fact that cost
(reward) function $C_{i}^{\text{'}}$ is bounded, we have 
\begin{equation}
\bigl\Vert\mathbb{E}\bigl[\boldsymbol{M}^{\boldsymbol{m}_{t'}}\bigr]\bigl\Vert_{2}\leq a_{11}m_{Q}^{1/2},\label{eq:M_bound}
\end{equation}
with probability at least $1-L^{2}\mathrm{exp}\bigl\{-a_{8}m_{Q}R_{\boldsymbol{\omega}}^{2/3}L\bigr\}$.
Further, by substituting (\ref{eq:inner product term}) and (\ref{eq:M_bound})
into (\ref{eq:TD-learning decompose}), and rearranging the terms,
we can obtain that
\begin{align}
 & \bigl\Vert\mathbb{E}\bigl[\boldsymbol{m}_{t'}\bigr]-\dot{\boldsymbol{\omega}}_{t}\bigl\Vert_{2}^{2}\label{eq:bias1-4}\\
\leq & \frac{\bigl\Vert\mathbb{E}\bigl[\boldsymbol{m}_{t'}\bigr]-\dot{\boldsymbol{\omega}}_{t}\bigr\Vert_{2}^{2}-\bigl\Vert\mathbb{E}\bigl[\boldsymbol{m}_{t'+1}\bigr]-\dot{\boldsymbol{\omega}}_{t}\bigr\Vert_{2}^{2}}{\eta_{t'}\varsigma}+\frac{a_{11}^{2}m_{Q}\eta_{t'}}{\varsigma}.\nonumber 
\end{align}
Taking summation on both sides of (\ref{eq:bias1-4}), we obtain
\begin{align}
 & \stackrel[t'=n_{t}+1]{t}{\sum}\eta_{t'}\bigl\Vert\mathbb{E}\bigl[\boldsymbol{m}_{t'}\bigr]-\dot{\boldsymbol{\omega}}_{t}\bigl\Vert_{2}^{2}\nonumber \\
\leq & \frac{1}{\varsigma}\bigl\Vert\mathbb{E}\bigl[\boldsymbol{m}_{n_{t}+1}\bigr]-\dot{\boldsymbol{\omega}}_{t}\bigr\Vert_{2}^{2}+\frac{1}{\varsigma}a_{11}^{2}m_{Q}\eta_{n_{t}}^{2}t^{\kappa_{5}}\nonumber \\
\leq & O\bigl(\frac{1}{\varsigma m_{Q}}+\frac{a_{11}^{2}m_{Q}\eta_{n_{t}}^{2}t^{\kappa_{5}}}{\varsigma}\bigr),\label{eq:bias1-5}
\end{align}
where recall that $n_{t}=t-t^{\kappa_{5}}$. Finally, plugging (\ref{eq:bias1-5})
into the second term of (\ref{eq:bias1-2})-a, we can derive that{\small{}
\begin{align}
\frac{\gamma_{n_{t}}}{\eta_{n_{t}}} & \stackrel[t'=n_{t}+1]{t}{\sum}\eta_{t'}\bigl\Vert\mathbb{E}\bigl[\boldsymbol{m}_{t'}\bigr]-\dot{\boldsymbol{\omega}}_{t}\bigl\Vert_{2}^{2}\leq O\Bigl(\frac{\gamma_{n_{t}}+m_{Q}^{2}\gamma_{n_{t}}\eta_{n_{t}}^{2}t^{\kappa_{5}}}{m_{Q}\eta_{n_{t}}}\Bigr).\label{eq:bias1-6}
\end{align}
}Combining (\ref{eq:bias1-2}), (\ref{eq:bias1-3}), and (\ref{eq:bias1-6}),
we have{\small{}
\begin{align}
\bigl\Vert\mathbb{E}\bigl[\bar{\boldsymbol{m}}_{t}\bigr]-\dot{\boldsymbol{\omega}}_{t}\bigl\Vert_{2}^{2}\leq & O\Bigl(\frac{\gamma_{n_{t}}+m_{Q}^{2}\gamma_{n_{t}}\eta_{n_{t}}^{2}t^{\kappa_{5}}}{m_{Q}\eta_{n_{t}}}+\frac{\bigl(1-\gamma_{t}\bigr)^{t^{\kappa_{5}}}}{m_{Q}\gamma_{t}}\Bigr).\label{eq:bias2}
\end{align}
}{\small\par}

\paragraph{bias 3}

Together with (\ref{eq:w^_}) and (\ref{eq:m^_}), we obtain{\small{}
\begin{align}
 & \bigl\Vert\mathbb{E}\bigl[\bar{\boldsymbol{m}}_{t}\bigr]-\mathbb{E}\bigl[\bar{\boldsymbol{\omega}}_{t}\bigr]\bigr\Vert_{2}\label{eq:bias2-1}\\
\leq & \stackrel[t'=n_{t}+1]{t}{\sum}\stackrel[j=t'+1]{t}{\prod}\bigl(1-\gamma_{j}\bigr)\gamma_{t'}\bigl\Vert\mathbb{E}\bigl[\boldsymbol{m}_{t'}\bigr]-\mathbb{E}\bigl[\boldsymbol{\omega}_{t'}\bigr]\bigr\Vert_{2}\nonumber \\
\leq & \stackrel[t'=n_{t}+1]{t}{\sum}\bigl(1-\gamma_{t}\bigr)^{t-t'}\gamma_{t'}e_{n_{t}}^{b}=O\bigl(e_{n_{t}}^{b}\frac{\gamma_{n_{t}}}{\gamma_{t}}\bigr),\forall i,\nonumber 
\end{align}
}where $e_{n_{t}}^{b}=\mathrm{max}_{t'=n_{t}+1,\ldots,t}\left\{ \bigl\Vert\mathbb{E}\bigl[\boldsymbol{m}_{t'}\bigr]-\mathbb{E}\bigl[\boldsymbol{\omega}_{t'}\bigr]\bigr\Vert_{2}\right\} $.
According to Assumption 2, we have $\gamma_{n_{t}}/\gamma_{t}<\infty$,
and thus $O\bigl(e_{n_{t}}^{b}\gamma_{n_{t}}/\gamma_{t}\bigr)=O\bigl(e_{n_{t}}^{b}\bigr)$.

Combining (\ref{eq:TD-learning}) and (\ref{eq:m_TD_learning}), we
have{\small{}
\begin{align}
 & \bigl\Vert\mathbb{E}\bigl[\boldsymbol{m}_{t'}\bigr]-\mathbb{E}\bigl[\boldsymbol{\omega}_{t'}\bigr]\bigr\Vert_{2}\label{eq:bias2-2}\\
\leq & \stackrel[j=n_{t}+1]{t'}{\sum}\eta_{j}\bigl\Vert\mathbb{E}\bigl[\boldsymbol{\Delta}^{\boldsymbol{\omega}_{j-1}}\bigr]-\mathbb{E}\bigl[\boldsymbol{M}^{\boldsymbol{m}_{j-1}}\bigr]\bigl\Vert_{2}\nonumber \\
= & \stackrel[j=n_{t}+1]{t'}{\sum}\eta_{j}\bigl\Vert\varint_{\boldsymbol{s}\in\mathcal{S}}\mathbf{P}\bigl(S_{j}=\mathrm{d}\boldsymbol{s}\bigr)\varint_{\boldsymbol{a}\in\mathcal{A}}\pi_{\boldsymbol{\theta}_{j}}\bigl(\mathrm{d}\boldsymbol{a}\mid\boldsymbol{s}\bigr)\boldsymbol{\Delta}^{\boldsymbol{\omega}_{j-1}}\nonumber \\
 & -\varint_{\tilde{\boldsymbol{s}}\in\mathcal{S}}\mathbf{P}_{\pi_{\boldsymbol{\theta}_{t}}}\varint_{\tilde{\boldsymbol{a}}\in\mathcal{A}}\pi_{\boldsymbol{\theta}_{t}}\bigl(\mathrm{d}\tilde{\boldsymbol{a}}\mid\tilde{\boldsymbol{s}}\bigr)\boldsymbol{M}^{\boldsymbol{m}_{j-1}}\bigl\Vert_{2}\nonumber \\
\overset{a}{\leq} & m_{Q}^{1/2}\stackrel[j=n_{t}+1]{t'}{\sum}\eta_{j}O\Bigl(\bigl\Vert\varint_{\boldsymbol{s}\in\mathcal{S}}\mathbf{P}\bigl(S_{j}=\mathrm{d}\boldsymbol{s}\bigr)\varint_{\boldsymbol{a}\in\mathcal{A}}\pi_{\boldsymbol{\theta}_{j}}\bigl(\mathrm{d}\boldsymbol{a}\mid\boldsymbol{s}\bigr)\nonumber \\
 & -\varint_{\tilde{\boldsymbol{s}}\in\mathcal{S}}\mathbf{P}_{\pi_{\boldsymbol{\theta}_{t}}}\varint_{\boldsymbol{a}\in\mathcal{A}}\pi_{\boldsymbol{\theta}_{j}}\bigl(\mathrm{d}\boldsymbol{a}\mid\boldsymbol{s}\bigr)\bigl\Vert_{2}+\bigl\Vert\varint_{\tilde{\boldsymbol{s}}\in\mathcal{S}}\mathbf{P}_{\pi_{\boldsymbol{\theta}_{t}}}\nonumber \\
 & \varint_{\boldsymbol{a}\in\mathcal{A}}\pi_{\boldsymbol{\theta}_{j}}\bigl(\mathrm{d}\boldsymbol{a}\mid\boldsymbol{s}\bigr)-\varint_{\tilde{\boldsymbol{s}}\in\mathcal{S}}\mathbf{P}_{\pi_{\boldsymbol{\theta}_{t}}}\varint_{\tilde{\boldsymbol{a}}\in\mathcal{A}}\pi_{\boldsymbol{\theta}_{t}}\bigl(\mathrm{d}\tilde{\boldsymbol{a}}\mid\tilde{\boldsymbol{s}}\bigr)\bigl\Vert_{2}\Bigr)\nonumber \\
\overset{b}{\leq} & m_{Q}^{1/2}\stackrel[j=n_{t}+1]{t'}{\sum}\eta_{j}O\bigl(\bigl\Vert\mathbf{P}\bigl(S_{j}\in\cdot\bigr)-\mathbf{P}_{\pi_{\boldsymbol{\theta}_{t}}}\bigr\Vert_{TV}+\bigl\Vert\boldsymbol{\theta}_{j}-\boldsymbol{\theta}_{t}\bigr\Vert_{2}\bigr),\nonumber 
\end{align}
}where (\ref{eq:bias2-2})-a uses the bounded property of $\boldsymbol{\Delta}^{\boldsymbol{\omega}_{j-1}}$
and $\boldsymbol{M}^{\boldsymbol{m}_{j-1}}$ and follows from triangle
inequality. Moreover, (\ref{eq:bias2-2})-b comes from that the policy
$\pi_{\boldsymbol{\theta}}$ follows the Lipschitz continuity over
$\boldsymbol{\theta}$. For the first term of (\ref{eq:bias2-2})-b,
we have{\small{}
\begin{align}
 & \bigl\Vert\mathbf{P}\bigl(S_{j}\in\cdot\bigr)-\mathbf{P}_{\pi_{\boldsymbol{\theta}_{t}}}\bigr\Vert_{TV}\label{eq:bias2-4}\\
= & \bigl\Vert\mathbf{P}\bigl(S_{j}\in\cdot\bigr)-\mathbf{P}_{\pi_{\boldsymbol{\theta}_{j-\tau_{j}}}}\bigr\Vert_{TV}+\bigl\Vert\mathbf{P}_{\pi_{\boldsymbol{\theta}_{j-\tau_{j}}}}-\mathbf{P}_{\pi_{\boldsymbol{\theta}_{t}}}\bigr\Vert_{TV}\nonumber \\
\overset{a}{\leq} & \lambda\rho^{\tau_{j}}+\bigl(\bigl\lceil\mathrm{log}{}_{\rho}\lambda^{-1}\bigr\rceil+\frac{1}{1-\rho}\bigr)O\bigl(\stackrel[k=j-\tau_{j}+1]{t}{\sum}\beta_{k}\bigr),\nonumber 
\end{align}
}where (\ref{eq:bias2-4})-a follows the ergodicity assumption and
the theoretical result of (35) and (36) in \cite{SCAOPO}, and we
let $\rho^{\tau_{j}}=O\left(\frac{1}{t}\right)$. Moreover, for the
second term of (\ref{eq:bias2-2})-b, we have{\small{}
\begin{equation}
\bigl\Vert\boldsymbol{\theta}_{j}-\boldsymbol{\theta}_{t}\bigr\Vert_{2}=O\bigl(\stackrel[k=j+1]{t}{\sum}\beta_{k}\bigr).\label{eq:bias2-5}
\end{equation}
}By plugging (\ref{eq:bias2-4}) and (\ref{eq:bias2-5}) into (\ref{eq:bias2-2}),
we obtain
\begin{align}
 & \bigl\Vert\mathbb{E}\bigl[\boldsymbol{m}_{t'}\bigr]-\mathbb{E}\bigl[\boldsymbol{\omega}_{t'}\bigr]\bigr\Vert_{2}\label{eq:bias2-6}\\
\leq & m_{Q}^{1/2}\stackrel[j=n_{t}+1]{t'}{\sum}O\Bigl(\eta_{j}\lambda\rho^{\tau_{j}}+\eta_{j}\stackrel[k=j-\tau_{j}+1]{t}{\sum}\beta_{k}\Bigr)\nonumber \\
\leq & O\Bigl(m_{Q}^{1/2}\eta_{n_{t}}t^{\kappa_{5}-1}+m_{Q}^{1/2}\eta_{n_{t}}\beta_{n_{t}}t^{2\kappa_{5}}\Bigr)\text{,}\nonumber 
\end{align}
Thus, we have $e_{n_{t}}^{b}\leq m_{Q}^{1/2}\eta_{n_{t}}O\bigl(t^{-1}\delta_{t}^{-1}+\beta_{n_{t}}\delta_{t}^{-2}\bigr)$,
and further have
\begin{equation}
\bigl\Vert\mathbb{E}\bigl[\bar{\boldsymbol{m}}_{t}\bigr]-\mathbb{E}\bigl[\bar{\boldsymbol{\omega}}_{t}\bigr]\bigr\Vert_{2}\leq m_{Q}^{1/2}\eta_{n_{t}}O\bigl(t^{\kappa_{5}-1}+\beta_{n_{t}}t^{2\kappa_{5}}\bigr).\label{eq:bias3}
\end{equation}

Combining (\ref{eq:divide-1}), (\ref{eq:bias1}), (\ref{eq:error1-1}),
(\ref{eq:divide-2}), (\ref{eq:bias2}), and (\ref{eq:bias3}), we
can finally obtain{\small{}
\begin{align}
 & \Bigl|\mathbb{E}\bigl[f\bigl(\bar{\boldsymbol{\omega}}_{t}^{i}\bigr)\bigr]-\hat{Q}_{i}^{\pi_{\boldsymbol{\theta}_{t}}}\Bigl|\leq O\Bigl(\frac{\bigl(1-\gamma_{t}\bigr)^{t^{\kappa_{5}}/2}}{\gamma_{t}^{1/2}}++m_{Q}\eta_{n_{t}}t^{\kappa_{5}-1}\nonumber \\
+ & m_{Q}\gamma_{n_{t}}^{1/2}\eta_{n_{t}}^{1/2}t^{\kappa_{5}/2}+\frac{\gamma_{n_{t}}^{1/2}}{\eta_{n_{t}}^{1/2}}+m_{Q}\eta_{n_{t}}\beta_{n_{t}}t^{2\kappa_{5}}+\epsilon_{m_{Q}}\Bigr),\label{eq:combine_bias1-1}
\end{align}
}with probability at least $1-L^{2}\mathrm{exp}\bigl\{-a_{9}m_{Q}R_{\boldsymbol{\omega}}^{2/3}L\bigr\}$.
Finally, by setting $\kappa_{5}=0.43$ in particular, we can finally
obtain the theoretical result in Lemma 1. This completes the proof.

\section{Proof of Lemma 2}

Our proof of Lemma 2 relies on a technical lemma \cite[Lemma 1]{Lemma3},
which is restated below for completeness.

\begin{lemma}\rm Let $(\Omega,\mathcal{G},\mathbb{P})$ denote a
probability space and let $\{\mathcal{G}_{t}\}$ denote an increasing
sequence of $\sigma$-field contained in $\mathcal{G}$. Let $\{z^{t}\}$,
$\{w^{t}\}$ be sequences of $\mathcal{G}_{t}$-measurable random
vectors satisfying the relations
\begin{align}
w^{t+1}= & \prod_{\mathcal{W}}(w^{t}+\alpha_{t}(\varrho^{t}-w^{t}))\\
\mathbb{E}[\varrho^{t}|\mathcal{G}_{t}]= & z^{t}+o^{t}\nonumber 
\end{align}
where $\alpha_{t}\geq0$ and the set $\mathcal{W}$ is convex and
closed, $\prod_{\mathcal{W}}(\cdot)$ denotes projection on $\mathcal{W}$.
Let\\
(a) all accumulation points of $\{w^{t}\}$ belong to $\mathcal{W}$
w.p.l.,\\
(b) there is a constant $C$ such that $\mathbb{E}[\Vert\varrho^{t}\Vert_{2}|\mathcal{G}_{t}]\leq C$,
$\forall t\geq0$,\\
(c) $\sum_{t=0}^{\infty}\mathbb{E}[(\alpha_{t})^{2}+\alpha_{t}\Vert o^{t}\Vert]\text{<}\infty$\\
(d) $\sum_{t=0}^{\infty}\alpha_{t}=\infty$, and (e) $\Vert w^{t+1}-w^{t}\Vert/\alpha_{t}\rightarrow0$
w.p.l.,\\
Then $z^{t}-w^{t}\rightarrow0\ w.p.1.$\end{lemma}

By Lemma 5, we can prove the asymptotic consistency of function values
(\ref{eq:JJ}) following the similar analyses in Appendix A-A of \cite{SCAOPO},
and we omit the proof due to the space limit. Then we give the proof
of (\ref{eq:gg}).

We first construct an auxiliary policy gradient estimate:
\begin{equation}
\nabla_{\boldsymbol{\theta}}\hat{J}_{i}\left(\boldsymbol{\theta}\right)=\mathbb{E}_{\sigma_{\pi_{\boldsymbol{\theta}}}}\left[\hat{Q}_{i}^{\pi_{\boldsymbol{\theta}}}\left(\boldsymbol{s},\boldsymbol{a}\right)\nabla_{\boldsymbol{\theta}}\textrm{log}\pi_{\boldsymbol{\theta}}\left(\boldsymbol{a}\mid\boldsymbol{s}\right)\right],\forall i,\label{eq:auxiliary policy gradient}
\end{equation}
Note that the only difference between (\ref{eq:real policy gradient})
and (\ref{eq:auxiliary policy gradient}) is that we replace the exact
Q-value $Q_{i}^{\pi_{\boldsymbol{\theta}_{t}}}$ with the approximate
Q-functions $\hat{Q}_{i}^{\pi_{\boldsymbol{\theta}_{t}}}$. Then the
asymptotic consistency of (\ref{eq:gg}) can be decomposed into the
following two steps:
\begin{align}
\mathrm{Step}1: & \underset{t\rightarrow\infty}{\mathrm{lim}}\bigl\Vert\nabla_{\boldsymbol{\theta}}\hat{J}_{i}\left(\boldsymbol{\theta}\right)-\nabla_{\boldsymbol{\theta}}J_{i}\left(\boldsymbol{\theta}_{t}\right)\bigr\Vert_{2}=0\label{eq:lim_g_step1}\\
\mathrm{Step}2: & \underset{t\rightarrow\infty}{\mathrm{lim}}\bigl\Vert\hat{\boldsymbol{g}}_{i}^{t}-\nabla_{\boldsymbol{\theta}}\hat{J}_{i}\left(\boldsymbol{\theta}\right)\bigr\Vert_{2}\leq\epsilon_{m_{Q}}\label{eq:lim_g_step2}
\end{align}

\paragraph{Step 1}

Combining (\ref{eq:real policy gradient}) with (\ref{eq:auxiliary policy gradient}),
and together with the regularity conditions in Assumption 1, i.e.,
$C_{i}^{'}$ and $\nabla_{\boldsymbol{\theta}}\textrm{log}\pi_{\boldsymbol{\theta}}$
are bounded, we have
\begin{align}
\bigl\Vert\nabla_{\boldsymbol{\theta}}\hat{J}_{i}\left(\boldsymbol{\theta}_{t}\right)-\nabla J_{i}\left(\boldsymbol{\theta}_{t}\right)\bigr\Vert_{2}= & O\bigl(\bigl\Vert\hat{J}_{i}^{t}-J_{i}\left(\boldsymbol{\theta}_{t}\right)\bigr\Vert_{2}\bigr).
\end{align}
Recalling (\ref{eq:JJ}), we can complete the proof of step 1.

\paragraph{Step 2}

Since the step size $\left\{ \alpha_{t}\right\} $ follows Assumption
3, and $C_{i}(\boldsymbol{s},\boldsymbol{a}),\forall i$ is bounded,
it's easy to prove that the conditions (a), (b), and (d) in Lemma
4 are satisfied. Now, we are ready to prove the technical condition
(c).

Together with the definition of real gradient $\nabla J_{i}\left(\boldsymbol{\theta}_{t}\right)$
in (\ref{eq:real policy gradient}), we obtain the stochastic policy
gradient error:{\small{}
\begin{align}
 & \left\Vert o^{t}\right\Vert _{2}=\bigl\Vert\tilde{g}_{i}^{t+1}-\nabla_{\boldsymbol{\theta}}J_{i}\left(\boldsymbol{\theta}_{t}\right)\bigr\Vert_{2}\label{eq:stochastic bias}\\
\overset{a}{=} & \frac{1}{T_{t}}\stackrel[l=1]{T_{t}}{\sum}O\bigl(\bigl\Vert\boldsymbol{\theta}_{t}-\boldsymbol{\theta}_{t-T_{t}+l}\bigr\Vert_{2}+\bigl\Vert\mathbf{P}\bigl(S_{t-T_{t}+l}\in\cdot\bigr)-\mathbf{P}_{\pi_{\boldsymbol{\theta}_{t}}}\bigr\Vert_{TV}\nonumber \\
 & +\bigl|\mathbb{E}\bigl[\hat{f}\bigl(\bar{\boldsymbol{\omega}}_{t}^{i}\bigr)\bigr]-\hat{Q}_{i}^{\pi_{\boldsymbol{\theta}_{t}}}\bigr|\bigr),\nonumber 
\end{align}
}where (\ref{eq:stochastic bias})-a follows from similar tricks as
in (\ref{eq:bias2-2}), triangle inequality, and the regularity conditions
in Assumption 1, i.e., the policy $\pi_{\boldsymbol{\theta}}$ follows
the Lipschitz continuity over $\boldsymbol{\theta}$, and $C_{i}^{'}$,
the output of DNNs and $\nabla_{\boldsymbol{\theta}}\textrm{log}\pi_{\boldsymbol{\theta}}$
are bounded. Then, plugging (\ref{eq:bias2-4}), (\ref{eq:bias2-5})
and Lemma 1 into (\ref{eq:stochastic bias}), we finally obtain
\begin{align}
\left\Vert o^{t}\right\Vert _{2}= & O\bigl(\epsilon_{m_{Q}}+t^{-1}+T_{t}\beta_{t-T_{t}}+\frac{\bigl(1-\gamma_{t}\bigr)^{t^{\kappa_{5}}/2}}{\gamma_{t}^{1/2}}+\frac{\gamma_{n_{t}}^{1/2}}{\eta_{n_{t}}^{1/2}}\nonumber \\
+m_{Q}\gamma_{n_{t}}^{1/2} & \eta_{n_{t}}^{1/2}t^{\kappa_{5}/2}+m_{Q}\eta_{n_{t}}t^{\kappa_{5}-1}+m_{Q}\eta_{n_{t}}\beta_{n_{t}}t^{2\kappa_{5}}\bigr),
\end{align}
where we set $\rho^{\tau_{t}}=O\bigl(\frac{1}{t}\bigr)$, which means
$\tau_{t}=O\bigl(\mathrm{log}\thinspace t\bigr)$. According to Assumption
2, it's easy to prove that condition (c) is held. For the condition
(e), we have
\begin{align}
 & \bigl\Vert\nabla_{\boldsymbol{\theta}}\hat{J}_{i}\left(\boldsymbol{\theta}_{t+1}\right)-\nabla_{\boldsymbol{\theta}}\hat{J}_{i}\left(\boldsymbol{\theta}_{t}\right)\bigr\Vert_{2}\\
= & O\bigl(\bigl\Vert\boldsymbol{\theta}_{t+1}-\boldsymbol{\theta}_{t}\bigr\Vert_{2}+\bigl\Vert\mathbf{P}_{\pi_{\boldsymbol{\theta}_{t+1}}}-\mathbf{P}_{\pi_{\boldsymbol{\theta}_{t}}}\bigr\Vert_{TV}\bigr)=O\bigl(\beta_{t}\bigr)\nonumber 
\end{align}
It can be seen from Assumption 2 that technical condition (e) is also
satisfied. This completes the proof of step 2.

\end{appendices}

\bibliographystyle{IEEEtran}
\bibliography{RLreferences}

\begin{thebibliography}{10}
\providecommand{\url}[1]{#1}
\csname url@samestyle\endcsname
\providecommand{\newblock}{\relax}
\providecommand{\bibinfo}[2]{#2}
\providecommand{\BIBentrySTDinterwordspacing}{\spaceskip=0pt\relax}
\providecommand{\BIBentryALTinterwordstretchfactor}{4}
\providecommand{\BIBentryALTinterwordspacing}{\spaceskip=\fontdimen2\font plus
\BIBentryALTinterwordstretchfactor\fontdimen3\font minus
  \fontdimen4\font\relax}
\providecommand{\BIBforeignlanguage}[2]{{%
\expandafter\ifx\csname l@#1\endcsname\relax
\typeout{** WARNING: IEEEtran.bst: No hyphenation pattern has been}%
\typeout{** loaded for the language `#1'. Using the pattern for}%
\typeout{** the default language instead.}%
\else
\language=\csname l@#1\endcsname
\fi
#2}}
\providecommand{\BIBdecl}{\relax}
\BIBdecl

\bibitem{DRL}
K.~Arulkumaran, M.~P. Deisenroth, M.~Brundage, and A.~A. Bharath, ``Deep
  reinforcement learning: A brief survey,'' \emph{IEEE Trans. Signal Process.},
  vol.~34, no.~6, pp. 26--38, 2017.

\bibitem{Go}
D.~Silver, A.~Huang, C.~J. Maddison, A.~Guez, L.~Sifre, G.~Van Den~Driessche,
  J.~Schrittwieser, I.~Antonoglou, V.~Panneershelvam, M.~Lanctot \emph{et~al.},
  ``Mastering the game of {G}o with deep neural networks and tree search,''
  \emph{Nature}, vol. 529, no. 7587, pp. 484--489, 2016.

\bibitem{Atari}
V.~Mnih, K.~Kavukcuoglu, D.~Silver, A.~Graves, I.~Antonoglou, D.~Wierstra, and
  M.~Riedmiller, ``Playing {A}tari with deep reinforcement learning,'' 2013.

\bibitem{medical}
T.~W. Bickmore, H.~Trinh, S.~Olafsson, T.~K. O'Leary, R.~Asadi, N.~M. Rickles,
  and R.~Cruz, ``Patient and consumer safety risks when using conversational
  assistants for medical information: an observational study of {S}iri,
  {A}lexa, and {G}oogle assistant,'' \emph{J Med Internet Res}, vol.~20, no.~9,
  p. e11510, 2018.

\bibitem{robotlocomotion}
J.~Schulman, S.~Levine, P.~Abbeel, M.~Jordan, and P.~Moritz, ``Trust region
  policy optimization,'' in \emph{ICML}.\hskip 1em plus 0.5em minus 0.4em\relax
  PMLR, 2015, pp. 1889--1897.

\bibitem{RRM}
C.-X. Wang, J.~Wang, S.~Hu, Z.~H. Jiang, J.~Tao, and F.~Yan, ``Key technologies
  in 6{G} terahertz wireless communication systems: A survey,'' \emph{IEEE Veh.
  Technol. Mag.}, vol.~16, no.~4, pp. 27--37, 2021.

\bibitem{SCAOPO}
C.~Tian, A.~Liu, G.~Huang, and W.~Luo, ``Successive convex approximation based
  off-policy optimization for constrained reinforcement learning,'' \emph{IEEE
  Trans. Signal Process.}, vol.~70, pp. 1609--1624, 2022.

\bibitem{reviewer1_1_3}
J.~Zhang, A.~S. Bedi, M.~Wang, and A.~Koppel, ``Cautious reinforcement learning
  via distributional risk in the dual domain,'' \emph{IEEE Journal on Selected
  Areas in Information Theory}, vol.~2, no.~2, pp. 611--626, 2021.

\bibitem{reviewer1_1_1}
J.~Zhang, A.~Koppel, A.~S. Bedi, C.~Szepesvari, and M.~Wang, ``Variational
  policy gradient method for reinforcement learning with general utilities,''
  in \emph{Advances in Neural Information Processing Systems}, H.~Larochelle,
  M.~Ranzato, R.~Hadsell, M.~Balcan, and H.~Lin, Eds., vol.~33.\hskip 1em plus
  0.5em minus 0.4em\relax Curran Associates, Inc., 2020, pp. 4572--4583.

\bibitem{reviewer1_4_6}
D.~Ding, K.~Zhang, T.~Basar, and M.~R. Jovanovic, ``Convergence and optimality
  of policy gradient primal-dual method for constrained markov decision
  processes,'' in \emph{2022 American Control Conference (ACC)}, 2022, pp.
  2851--2856.

\bibitem{primal_dual_conv1}
Y.~Chen, J.~Dong, and Z.~Wang, ``A primal-dual approach to constrained markov
  decision processes,'' \emph{arXiv preprint arXiv:2101.10895}, 2021.

\bibitem{primal_dual_conv2}
D.~Ding, K.~Zhang, T.~Basar, and M.~Jovanovic, ``Natural policy gradient
  primal-dual method for constrained markov decision processes,''
  \emph{Advances in Neural Information Processing Systems}, vol.~33, pp.
  8378--8390, 2020.

\bibitem{SCAOPO17}
A.~Stooke, J.~Achiam, and P.~Abbeel, ``Responsive safety in reinforcement
  learning by pid lagrangian methods,'' in \emph{International Conference on
  Machine Learning}.\hskip 1em plus 0.5em minus 0.4em\relax PMLR, 2020, pp.
  9133--9143.

\bibitem{SCAOPO18}
D.~Ding, X.~Wei, Z.~Yang, Z.~Wang, and M.~Jovanovic, ``Provably efficient safe
  exploration via primal-dual policy optimization,'' in \emph{International
  Conference on Artificial Intelligence and Statistics}.\hskip 1em plus 0.5em
  minus 0.4em\relax PMLR, 2021, pp. 3304--3312.

\bibitem{reviewer1_4_8}
A.~Muller, P.~Alatur, G.~Ramponi, and N.~He, ``Cancellation-free regret bounds
  for lagrangian approaches in constrained markov decision processes,'' 2023.

\bibitem{reviewer1_4_7}
D.~Ding, C.-Y. Wei, K.~Zhang, and A.~Ribeiro, ``Last-iterate convergent policy
  gradient primal-dual methods for constrained mdps,'' 2023.

\bibitem{PPOLagTRPOLag}
J.~Achiam and D.~Amodei, ``Benchmarking safe exploration in deep reinforcement
  learning,'' 2019.

\bibitem{reviewer1_4_5}
M.~Gaur, A.~S. Bedi, D.~Wang, and V.~Aggarwal, ``On the global convergence of
  natural actor-critic with two-layer neural network parametrization,''
  \emph{arXiv preprint arXiv:2306.10486}, 2023.

\bibitem{14in1_4_5}
Z.~Yang, Y.~Chen, M.~Hong, and Z.~Wang, ``Provably global convergence of
  actor-critic: A case for linear quadratic regulator with ergodic cost,''
  \emph{Advances in neural information processing systems}, vol.~32, 2019.

\bibitem{2019b}
B.~Liu, Q.~Cai, Z.~Yang, and Z.~Wang, ``Neural proximal/trust region policy
  optimization attains globally optimal policy,'' \emph{arXiv preprint
  arXiv:1906.10306}, 2019.

\bibitem{CRLcategory}
Q.~Liang, F.~Que, and E.~Modiano, ``Accelerated primal-dual policy optimization
  for safe reinforcement learning,'' \emph{arXiv preprint arXiv:1802.06480},
  2018.

\bibitem{reviewer1_1_2}
S.~Paternain, L.~Chamon, M.~Calvo-Fullana, and A.~Ribeiro, ``Constrained
  reinforcement learning has zero duality gap,'' in \emph{Advances in Neural
  Information Processing Systems}, H.~Wallach, H.~Larochelle, A.~Beygelzimer,
  F.~d\textquotesingle Alch\'{e}-Buc, E.~Fox, and R.~Garnett, Eds.,
  vol.~32.\hskip 1em plus 0.5em minus 0.4em\relax Curran Associates, Inc.,
  2019.

\bibitem{CPO}
J.~Achiam, D.~Held, A.~Tamar, and P.~Abbeel, ``Constrained policy
  optimization,'' in \emph{ICML}.\hskip 1em plus 0.5em minus 0.4em\relax PMLR,
  2017, pp. 22--31.

\bibitem{boundedJ2}
S.~Zou, T.~Xu, and Y.~Liang, ``Finite-sample analysis for sarsa with linear
  function approximation,'' \emph{Adv. Neural Inf. Process. Syst.}, vol.~32,
  2019.

\bibitem{offlineCRLrecommendedbyreviewer2}
Z.~Liu, Z.~Guo, Y.~Yao, Z.~Cen, W.~Yu, T.~Zhang, and D.~Zhao, ``Constrained
  decision transformer for offline safe reinforcement learning,'' in
  \emph{International Conference on Machine Learning}.\hskip 1em plus 0.5em
  minus 0.4em\relax PMLR, 2023, pp. 21\,611--21\,630.

\bibitem{reviewer1_4_3}
Y.~F. Wu, W.~ZHANG, P.~Xu, and Q.~Gu, ``A finite-time analysis of two
  time-scale actor-critic methods,'' in \emph{Advances in Neural Information
  Processing Systems}, H.~Larochelle, M.~Ranzato, R.~Hadsell, M.~Balcan, and
  H.~Lin, Eds., vol.~33.\hskip 1em plus 0.5em minus 0.4em\relax Curran
  Associates, Inc., 2020, pp. 17\,617--17\,628.

\bibitem{9in1_4_4}
X.~Chen, J.~Duan, Y.~Liang, and L.~Zhao, ``Global convergence of two-timescale
  actor-critic for solving linear quadratic regulator,'' in \emph{Proceedings
  of the AAAI Conference on Artificial Intelligence}, vol.~37, no.~6, 2023, pp.
  7087--7095.

\bibitem{11in1_4_4}
T.~Xu, Z.~Wang, and Y.~Liang, ``Non-asymptotic convergence analysis of two
  time-scale (natural) actor-critic algorithms,'' \emph{arXiv preprint
  arXiv:2005.03557}, 2020.

\bibitem{reviewer1_4_4}
X.~Chen and L.~Zhao, ``Finite-time analysis of single-timescale actor-critic,''
  2023.

\bibitem{URLLC2}
A.~Destounis and G.~S. Paschos, ``Complexity of {URLLC} scheduling and
  efficient approximation schemes,'' in \emph{Proc. Int. Symp. Modeling and
  Optim. Mobile, Ad Hoc, and Wireless Netw. (WiOPT)}, 2019, pp. 1--8.

\bibitem{XR}
X.~Zhao, Y.-J.~A. Zhang, M.~Wang, X.~Chen, and Y.~Li, ``Online multi-user
  scheduling for xr transmissions with hard-latency constraint: Performance
  analysis and practical design,'' \emph{IEEE Trans. Commun.}, 2024.

\bibitem{URLLC3}
H.~Ren, C.~Pan, Y.~Deng, M.~Elkashlan, and A.~Nallanathan, ``Resource
  allocation for secure {URLLC} in mission-critical {I}o{T} scenarios,''
  \emph{IEEE Trans. Commun.}, vol.~68, no.~9, pp. 5793--5807, 2020.

\bibitem{RZF}
C.~B. Peel, B.~M. Hochwald, and A.~L. Swindlehurst, ``A vector-perturbation
  technique for near-capacity multiantenna multiuser communication-part {I}:
  channel inversion and regularization,'' \emph{IEEE Trans. Commun.}, vol.~53,
  no.~1, pp. 195--202, 2005.

\bibitem{LQRimportance}
B.~Recht, ``A tour of reinforcement learning: The view from continuous
  control,'' \emph{Annu. Rev. Control, Robot., Auton. Syst.}, vol.~2, pp.
  253--279, 2019.

\bibitem{LQR}
M.~Fazel, R.~Ge, S.~Kakade, and M.~Mesbahi, ``Global convergence of policy
  gradient methods for the linear quadratic regulator,'' in \emph{ICML}.\hskip
  1em plus 0.5em minus 0.4em\relax PMLR, 2018, pp. 1467--1476.

\bibitem{average1}
C.-Y. Wei, M.~J. Jahromi, H.~Luo, H.~Sharma, and R.~Jain, ``Model-free
  reinforcement learning in infinite-horizon average-reward markov decision
  processes,'' in \emph{International conference on machine learning}.\hskip
  1em plus 0.5em minus 0.4em\relax PMLR, 2020, pp. 10\,170--10\,180.

\bibitem{Cao2019a}
Y.~Cao and Q.~Gu, ``Generalization error bounds of gradient descent for
  learning over-parameterized deep relu networks,'' in \emph{Proc. AAAI Conf.
  Artif. Intell.}, vol.~34, no.~04, 2020, pp. 3349--3356.

\bibitem{Allen2019b}
Z.~Allen-Zhu, Y.~Li, and Z.~Song, ``A convergence theory for deep learning via
  over-parameterization,'' in \emph{ICML}.\hskip 1em plus 0.5em minus
  0.4em\relax PMLR, 2019, pp. 242--252.

\bibitem{Cao2019b}
Y.~Cao and Q.~Gu, ``Generalization bounds of stochastic gradient descent for
  wide and deep neural networks,'' \emph{Proc. Adv. Neural Inf. Process.
  Syst.}, vol.~32, 2019.

\bibitem{Gaussianpolicy}
R.~S. Sutton and A.~G. Barto, \emph{Reinforcement learning: An
  introduction}.\hskip 1em plus 0.5em minus 0.4em\relax MIT press, 2018.

\bibitem{DQlearning}
P.~Xu and Q.~Gu, ``A finite-time analysis of {Q}-learning with neural network
  function approximation,'' in \emph{ICML}.\hskip 1em plus 0.5em minus
  0.4em\relax PMLR, 2020, pp. 10\,555--10\,565.

\bibitem{linnerAC}
S.~Qiu, Z.~Yang, J.~Ye, and Z.~Wang, ``On finite-time convergence of
  actor-critic algorithm,'' \emph{IEEE J. Sel. Areas Inf. Theory}, vol.~2,
  no.~2, pp. 652--664, 2021.

\bibitem{gradient0clipping}
A.~Koloskova, H.~Hendrikx, and S.~U. Stich, ``Revisiting gradient clipping:
  Stochastic bias and tight convergence guarantees,'' in \emph{Proceedings of
  the 40th International Conference on Machine Learning}, ser. Proceedings of
  Machine Learning Research, vol. 202.\hskip 1em plus 0.5em minus 0.4em\relax
  PMLR, 23--29 Jul 2023, pp. 17\,343--17\,363.

\bibitem{gradient1clipping}
J.~Qian, Y.~Wu, B.~Zhuang, S.~Wang, and J.~Xiao, ``Understanding gradient
  clipping in incremental gradient methods,'' in \emph{International Conference
  on Artificial Intelligence and Statistics}.\hskip 1em plus 0.5em minus
  0.4em\relax PMLR, 2021, pp. 1504--1512.

\bibitem{initialization}
X.~Glorot and Y.~Bengio, ``Understanding the difficulty of training deep
  feedforward neural networks,'' in \emph{Proceedings of the thirteenth
  international conference on artificial intelligence and statistics}.\hskip
  1em plus 0.5em minus 0.4em\relax JMLR Workshop and Conference Proceedings,
  2010, pp. 249--256.

\bibitem{RLintroduction}
R.~S. Sutton and A.~G. Barto, \emph{Reinforcement learning: An
  introduction}.\hskip 1em plus 0.5em minus 0.4em\relax MIT press, 2018.

\bibitem{CSSCA}
A.~Liu, V.~K.~N. Lau, and B.~Kananian, ``Stochastic successive convex
  approximation for non-convex constrained stochastic optimization,''
  \emph{IEEE Trans. Signal Process.}, vol.~67, no.~16, pp. 4189--4203, 2019.

\bibitem{Assumption31}
J.~Fan, Z.~Wang, Y.~Xie, and Z.~Yang, ``A theoretical analysis of deep
  {Q}-learning,'' in \emph{Proc. Learn. Dyn. Control}.\hskip 1em plus 0.5em
  minus 0.4em\relax PMLR, 2020, pp. 486--489.

\bibitem{Assumption32}
S.~Tosatto, M.~Pirotta, C.~d'Eramo, and M.~Restelli, ``Boosted fitted
  {Q}-iteration,'' in \emph{ICML}.\hskip 1em plus 0.5em minus 0.4em\relax PMLR,
  2017, pp. 3434--3443.

\bibitem{Yangrui}
A.~Liu, R.~Yang, T.~Q.~S. Quek, and M.-J. Zhao, ``Two-stage stochastic
  optimization via primal-dual decomposition and deep unrolling,'' \emph{IEEE
  Trans. Signal Process.}, vol.~69, pp. 3000--3015, 2021.

\bibitem{CLQset}
M.~Yu, Z.~Yang, M.~Kolar, and Z.~Wang, ``Convergent policy optimization for
  safe reinforcement learning,'' \emph{Adv. Neural Inf. Process. Syst.},
  vol.~32, 2019.

\bibitem{Lemma3}
A.~Ruszczy{\'n}ski, ``Feasible direction methods for stochastic programming
  problems,'' \emph{Math. Program.}, vol.~19, pp. 220--229, 1980.

\end{thebibliography}

\end{document}